\documentclass{article}
\usepackage[T1]{fontenc}
\usepackage{inconsolata}
\usepackage{tikz}
\usepackage{tikzpagenodes}
\usepackage{eso-pic}
\usepackage{etoolbox}       %
\usepackage{tabularx}       %

\usepackage{graphicx}  %

\usepackage[parfill]{parskip}
\usepackage{mathtools}  %
\usepackage{xcolor}         %
\usepackage{float}  %
\usepackage{listings}  %
\usepackage{pifont}

\usepackage{amsmath,amsthm,amssymb}
\usepackage{algorithm,algorithmic}
\allowdisplaybreaks

\newtoggle{arxiv}
\toggletrue{arxiv}

  \usepackage[citestyle=authoryear-comp,sorting=nyt,maxbibnames=99,backend=biber]{biblatex}
  \newcommand{\citep}{\parencite}
  \newcommand{\citet}{\textcite}
  \newlength{\defbaselineskip}
  \RequirePackage[T1]{fontenc}
  \RequirePackage[tt=false, type1=true]{libertine}
  \RequirePackage[libertine]{newtxmath}

\usepackage{bm}

\usepackage[inline]{enumitem}
\usepackage{booktabs}
\usepackage{subcaption}
\usepackage{multirow}
\usepackage{wrapfig}
\usepackage{nicematrix}

\usepackage[cachedir=.]{minted}

\usepackage[colorlinks=true,linkcolor=blue,citecolor=magenta,urlcolor=red]{hyperref}
\usepackage[capitalise,noabbrev]{cleveref}
\usepackage{xspace}

\usepackage{import}

\usepackage{pifont}

\definecolor{green}{RGB}{0,128,0}

\definecolor{yellow}{RGB}{255,200,18}

\newcommand{\cmark}{\textcolor{green!60!black}{\ding{51}}} 
\newcommand{\xmark}{\textcolor{red!70!black}{\ding{55}}}   

\newcommand{\diag}{\mathrm{diag}}

\newcommand{\vQ}{\mathbf{Q}}
\newcommand{\vK}{\mathbf{K}}
\newcommand{\vV}{\mathbf{V}}
\newcommand{\vdQ}{\mathbf{dQ}}
\newcommand{\vdK}{\mathbf{dK}}
\newcommand{\vdV}{\mathbf{dV}}
\newcommand{\vS}{\mathbf{S}}
\newcommand{\vdS}{\mathbf{dS}}
\newcommand{\vP}{\mathbf{P}}
\newcommand{\vdP}{\mathbf{dP}}

\newcommand{\vO}{\mathbf{O}}
\newcommand{\vdO}{\mathbf{dO}}
\newcommand{\vM}{\mathbf{M}}
\newcommand{\vdM}{\mathbf{dM}}

\title{Trainable Dynamic Mask Sparse Attention}
\usepackage{authblk}

\author[$^{13}$]{Jingze Shi}
\author[$^{13}$]{Yifan Wu}
\author[$^{13}$]{Yiran Peng}
\author[$^{13}$]{Bingheng Wu}
\author[$^2$]{Liangdong Wang}
\author[$^2$]{Guang Liu}
\author[$^1$]{Yuyu Luo\thanks{Corresponding author: Yuyu Luo (E-mail: yuyuluo@hkust-gz.edu.cn).}}
\affil[ ]{
    $^1$HKUST(GZ), 
    $^2$BAAI, 
    $^3$SmallDoges
}
\date{}

\begin{document}

\maketitle

\begin{abstract}
  The increasing demand for long-context modeling in large language models (LLMs) is bottlenecked by the quadratic complexity of the standard self-attention mechanism. The community has proposed sparse attention to mitigate this issue. However, position-aware sparse attention methods rely on static sparse structures that lack adaptability to diverse query contexts, while content-aware sparse attention methods depend on heuristic key-value selection, hindering full differentiability. We introduce a trainable dynamic mask sparse attention mechanism, a method that merges the advantages of both position-aware and content-aware approaches. Dynamic Mask Attention (DMA) achieves this through three key innovations: First, it leverages value vector representations to generate content-aware dynamic masks, enabling the model to adaptively identify and attend to critical information. Second, it computes position-aware sparse weights in a hardware-friendly manner, efficiently skipping unnecessary computational regions. Finally, we demonstrate that the introduced dynamic mask and sparse weights do not obstruct gradients, supporting end-to-end training. We have validated the performance of DMA through comprehensive experiments. A large body of experimental evidence shows that DMA consistently holds a Pareto advantage over state-of-the-art sparse attention baselines in tasks including scaling laws, multi-query associative recall, standard benchmarks, and needle in a haystack tests, while also delivering up to a 10x overall speedup. These results highlight its ability to effectively balance model efficiency with long-context modeling capabilities. Our computational kernel code is now open-source at \url{https://github.com/flash-algo/flash-sparse-attention} to encourage further research and application by the community.
\end{abstract}

\section{Introduction}
\label{sec:introduction}

Recent breakthroughs in large language models (LLMs) have yielded remarkable achievements in tasks requiring \textit{long-context reasoning}~\citep{snell2024tts}, such as deep reasoning~\citep{hf2025openr1}, codebase generation~\citep{zhang2024codeagent}, and multi-turn autonomous agents~\citep{park2023generative}. A key factor underpinning these successes is the ability to effectively model long-range dependencies, often spanning thousands of tokens~\citep{qwen32025, gemini2025, deepseekai2025deepseekr1incentivizingreasoningcapability}. However, the standard self-attention mechanism~\citep{vaswani2017attention} employed in Transformer architectures inherently exhibits quadratic computational complexity~\citep{zaheer2020big}, which severely restricts scalability to longer sequences. Consequently, designing attention mechanisms that enhance computational efficiency without sacrificing modeling accuracy has become a critical research frontier for advancing the capabilities of LLMs.

\paragraph{Limitations of Existing Methods.}
Current efficient attention strategies primarily leverage two types of sparsity: the \textbf{position sparsity}~\citep{martins2016softmax}, which facilitates the efficient computation of essential query-key pairs, and the \textbf{content sparsity}~\citep{ge2023model}, which enables the selective computation of relevant tokens.
The first category includes methods such as sliding window attention~\citep{beltagy2020longformerlongdocumenttransformer}, which employs static structures; multi-head latent attention~\citep{deepseekai2025deepseekv3technicalreport}, which uses low-rank approximations; and native sparse attention~\citep{yuan2025nativesparseattentionhardwarealigned}, which utilizes learnable compression weights. Although these approaches can achieve efficient long-context modeling, they often struggle to maintain their effectiveness.
The second category encompasses KV cache eviction methods~\citep{li2024snapkv, zhang2023h2o, zhou2024llm}; block-wise KV cache selection strategies that dynamically choose cache blocks based on relevance predictions~\citep{gao2024seerattention, tang2024quest, xiao2024infllm}; and filtering methods that employ sampling~\citep{chen2024magicpig}, clustering~\citep{liu2024clusterkv}, or hashing~\citep{desai2024hashattention}. Despite their conceptual appeal, these techniques often fail to realize their theoretical speedups in practical deployments due to the overhead from dynamic computations or inaccurate sparsification decisions.

\paragraph{Key Challenges.}
To overcome these limitations, an ideal sparse attention mechanism must simultaneously address two fundamental challenges: \textbf{leveraging position-aware sparsity for essential computations}~\citep{child2019generating} and \textbf{exploiting content-aware sparsity for selective computation}~\citep{dai2019transformerxl}. Meeting both requirements is crucial for achieving efficient and effective long-context reasoning and training in practice. However, existing methods still exhibit limitations, often facing a trade-off between efficiency and effectiveness. This dilemma highlights the urgent need for attention mechanisms that can preserve information integrity while achieving computational efficiency.

\begin{figure}[!t]
    \centering
    \includegraphics[width=\linewidth]{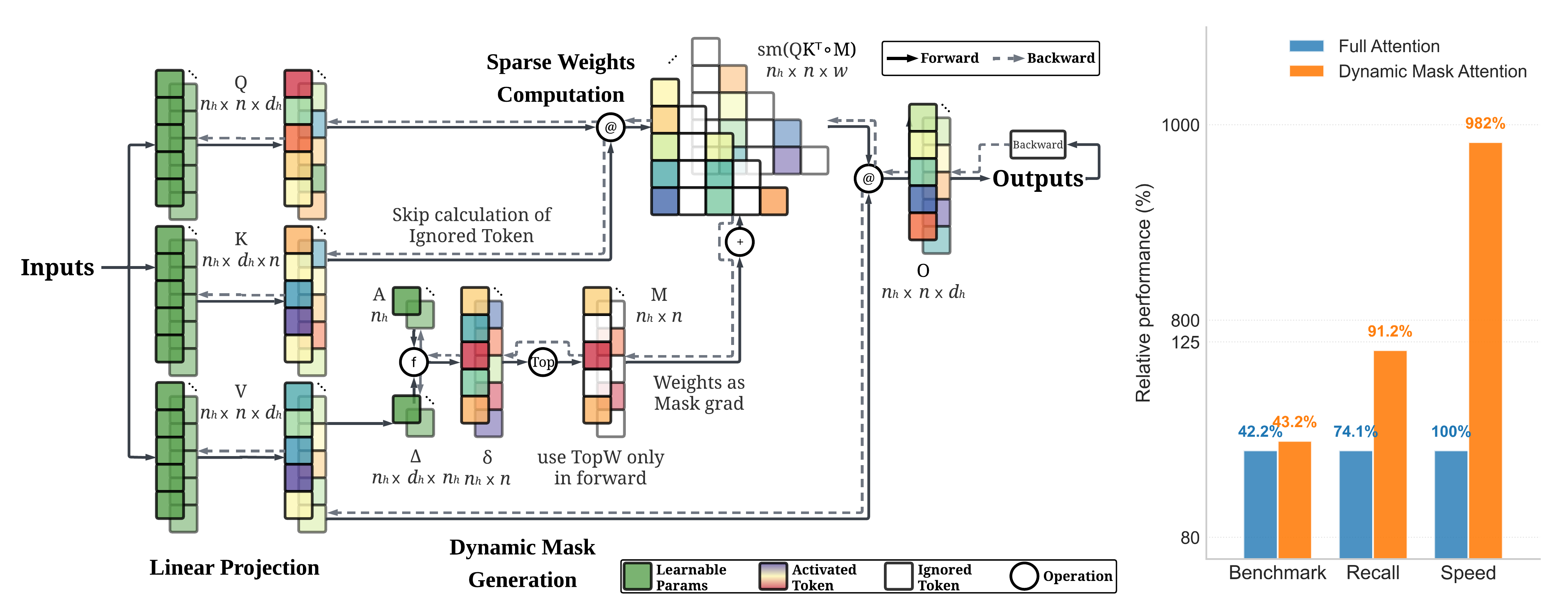}
    \vspace{-2em}
    \caption{
    \textbf{Workflow and Performance of Dynamic Mask Attention}.
    \textbf{Left:} Overall workflow of DMA. The first step projects the input into $Q$, $K$, and $V$. The second step generates content-aware dynamic masks. The third step computes sparse weights. Black solid arrows indicate the forward computation path, while gray dashed arrows represent the backward computation path. \textbf{Right:} Relative performance comparison between full attention and DMA on benchmark tests. DMA achieves higher recall rates and significantly faster speeds while maintaining competitive accuracy.
    }
    \label{fig:dma_flow}
\end{figure}
\vspace{-1em}
\begin{table}[!t]
    \centering
    \caption{
    \textbf{Comparison of Different Attention Variants}.
    Comparison of different Self-Attention mechanisms. $n$ denotes sequence length, $d_h$ represents head dimension, $w$ is window size, $d_c$ is compressed dimension, $B$ is compression block size, and $k$ is selection budget. Complexities focus on attention weight computation and memory requirements. Trainable indicates whether the sparsity pattern can be learned end-to-end.
    }
    \label{tab:attention_comparison_intro}
    \begin{tabular}{@{}lcccc@{}}
    \toprule
    \sc{Mechanism} & \sc{Comp. Complexity} & \sc{Mem. Complexity} & \sc{Sparsity} & \sc{Trainable} \\
    \midrule
    MHA & $O(n^2 d_h)$ & $O(n^2)$ & Static & \textcolor{orange}{\xmark} \\
    SWA & $O(n w d_h)$ & $O(n w)$ & Position-aware & \textcolor{orange}{\xmark} \\
    MLA & $O(n^2 d_c)$ & $O(n^2)$ & Low-rank Approx & \textcolor{green}{\cmark} \\
    NSA & $O(n^2 d_c / B + n k B d_h + n w d_h)$ & $O(n^2 / B + n k B)$ & Hybrid & \textcolor{green}{\cmark} \\
    H2O & $O(n k d_h)$ & $O(n k)$ & Content-aware & \textcolor{orange}{\xmark} \\
    InfLLM & $O(n k d_h)$ & $O(n k)$ & Content-aware & \textcolor{orange}{\xmark} \\
    Quest & $O(n k d_h)$ & $O(n k)$ & Content-aware & \textcolor{orange}{\xmark} \\
    DAM & $O(n k d_h)$ & $O(n k)$ & Content-aware & \textcolor{orange}{\xmark} \\
    \midrule
    \textbf{DMA} & $O(n w d_h)$ & $O(n w)$ & Content-Position Dual-aware & \textcolor{green}{\cmark} \\
    \bottomrule
    \end{tabular}
\end{table}

\begin{figure}[!t]
    \centering
    \includegraphics[width=\linewidth]{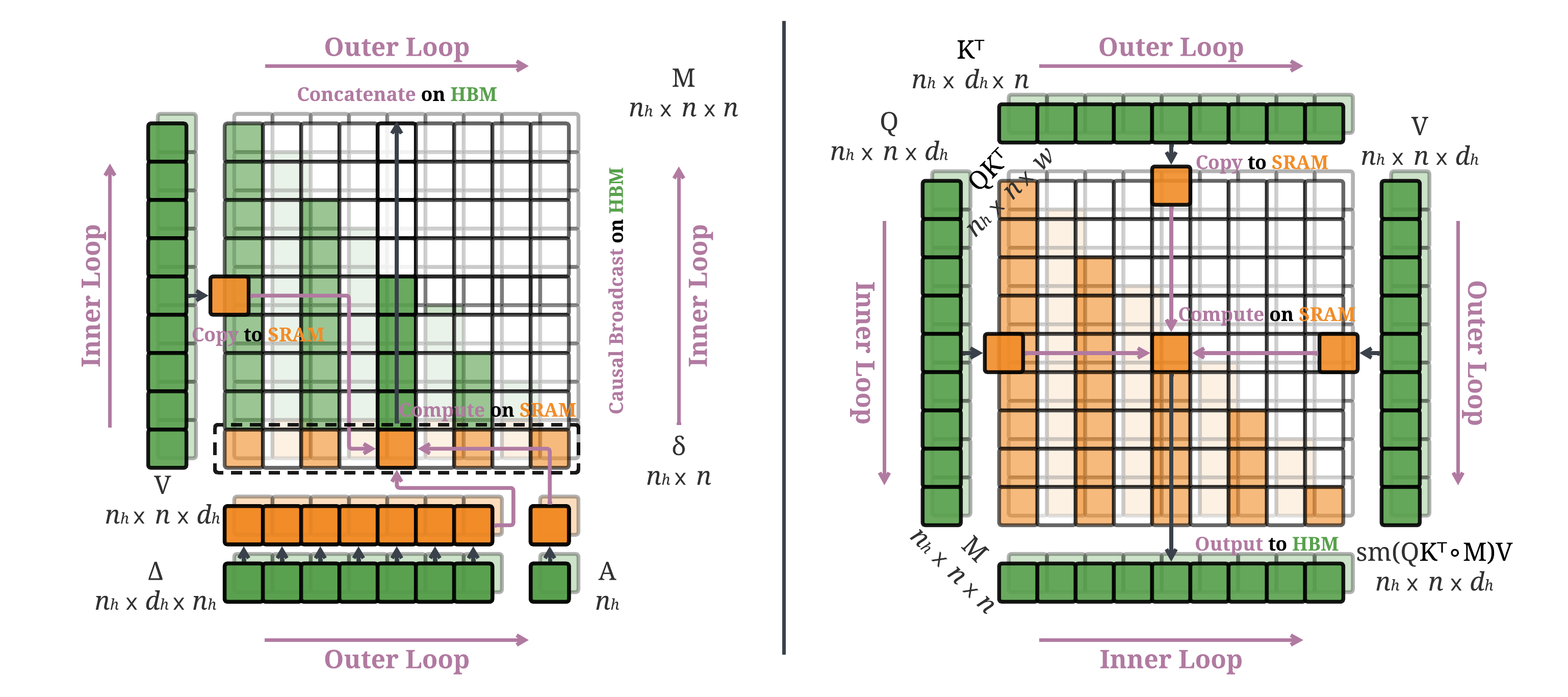}
    \caption{
    \textbf{Dynamic Mask Attention Architecture}.
    \textbf{Left}: \textbf{Content-Aware Mask Computation.} The mask computation part of dynamic mask attention. In the outer loop, the stride weight $\Delta$ and gate weight $A$ are loaded into high-speed SRAM, and in the inner loop, the zero-order hold method is used to loop through the $V$ blocks loaded into SRAM, sampling from it to generate content-aware $K$ masks. These masks are then causally broadcast to the length of $Q$ in HBM. Finally, in the outer loop, all mask blocks are concatenated to form the final content-aware sparse dynamic mask.
    \textbf{Right}: \textbf{Position-Aware Weights Computation.} The weight computation part of dynamic mask attention, where in the outer loop, the $K$ and $V$ blocks are looped and loaded into SRAM, and in the inner loop, the $Q$ blocks are accessed, loaded into SRAM, and the output of the attention weight computation is written back to HBM. If the current position of the $K$ block is designated as masked in the dynamic mask, the attention weight at that position is directly filled with 0, skipping the computation at that position, forming the final position-aware sparse attention weights.
    }
    \label{fig:dynamic_mask_attn}
\end{figure}

\paragraph{Our Method.}
In order to address these challenges and achieve efficient and effective sparse attention mechanisms, we ingeniously integrate the strengths of both strategies, attempting to strike a balance between the two, and propose Dynamic Mask Attention (DMA) to tackle the challenges of long-context modeling. As shown in Table~\ref{tab:attention_comparison_intro}, compared to other attention variants, DMA is a trainable content-position dual-aware sparse attention mechanism. As illustrated in Figure~\ref{fig:dma_flow}, it leverages two core innovations: \textbf{generating dynamic masks using content-aware sparsity} and \textbf{computing sparse weights using position-aware sparsity}, allowing the model to focus on relevant tokens while ignoring irrelevant ones. Furthermore, all computational operations are designed to be continuously differentiable, enabling end-to-end training of dynamic mask attention via gradient descent.

\paragraph{Kernel Design.}
We implement a dedicated CUDA kernel that merges the memory efficiency of FlashAttention~\citep{dao2022flashattention} with DMA's trainable sparsity, as illustrated in Figure~\ref{fig:dynamic_mask_attn}, enabling hardware-level skipping of masked regions without incurring additional redundant computations. The kernel natively supports attention masks and biases with batch, head, and query broadcasting, ensuring flexible integration with diverse Transformer architectures. A block-level reduction determines the skip logic: tiles corresponding to all-zero masks bypass both computation and memory access, reducing the effective complexity from $\mathcal{O}(n^2)$ to $\mathcal{O}(n \cdot w)$ for $w \ll n$. The forward and backward passes share a unified skip logic, fetching $K/V$ tiles only when necessary, thereby maintaining an $\mathcal{O}(n)$ memory footprint without materializing the full attention matrix. The backward pass incorporates a complete gradient chain with fused bias gradients, rendering the entire pipeline fully differentiable for end-to-end training. To maximize throughput, we employ shared memory aliasing, pipelined prefetching, and coalesced memory accesses to minimize bandwidth pressure and improve hardware occupancy. These design choices allow DMA to sustain high performance on extremely long contexts, such as 128K+ tokens, while preserving accuracy comparable to dense attention baselines.

\paragraph{Contributions.}
We make the following contributions: \textit{(i)} We introduce Dynamic Mask Attention (DMA), a trainable, content- and position-aware sparse attention mechanism that decouples content-driven dynamic mask generation from position-aware sparse weight computation, reducing the effective time complexity from $\mathcal{O}(n^{2})$ to $\mathcal{O}(n \cdot w)$ and the memory footprint to $\mathcal{O}(n \cdot w)$ for window size $w \ll n$. \textit{(ii)} We develop a fully differentiable CUDA kernel that fuses FlashAttention-style tiling with hardware-efficient mask skipping. The kernel supports batch/head/query broadcasting for masks and biases, employs block-level skip logic to avoid unnecessary computation and memory traffic, and maintains an $\mathcal{O}(n)$ memory footprint without materializing the full attention weight matrix. \textit{(iii)} We validate DMA across diverse application scenarios, showing consistent speedups over other optimized implementations, improved scaling laws~\citep{hoffmann2022empirical}, higher accuracy on multi-query associative recall~\citep{arora2024zoology}, competitive downstream benchmarks performance, and robust needle-in-a-haystack~\citep{gkamradt2023needleinahaystack} results.

\section{Rethinking Sparse Attention}
\label{sec:related_work}

Since the advent of the Transformer~\citep{vaswani2017attention}, the attention mechanism has become central to sequence modeling, yet its $O(n^2)$ computational and memory complexity remains a bottleneck for processing long sequences. To overcome this limitation, researchers have explored sparse attention from two core perspectives: \textbf{position-aware essential computation} and \textbf{content-aware selective computation}. The former reduces computational load through predefined sparse patterns, while the latter dynamically determines the scope of computation based on input content. This section reviews the evolution of sparse attention, analyzes the strengths and weaknesses of existing methods, and provides context for our proposed Dynamic Mask Attention.

\subsection{Position-Aware Essential Computation}

To achieve hardware efficiency, early sparse attention methods predominantly employed fixed sparse patterns, aiming to simplify computation through structured sparsity.

\textbf{Static Locality.}
Sliding Window Attention~\citep{beltagy2020longformerlongdocumenttransformer} confines computation to a local neighborhood for each token, reducing complexity to $O(n \cdot w)$, where $w$ is the window size. While simple and efficient, its fixed local window limits the model's ability to capture long-range dependencies. This limitation is particularly pronounced in tasks requiring information integration across window boundaries, such as long-form question answering or code analysis.

\textbf{Low-Rank Approximation.}
Methods like Multi-Head Latent Attention~\citep{deepseekai2025deepseekv3technicalreport} approximate the full attention matrix using low-rank decomposition to reduce computational and memory demands. While this approach is better at preserving global information than sliding window attention, it comes at the cost of precision loss due to information compression. Low-rank approximation can obscure fine-grained details crucial for specific tasks and, due to its global nature, cannot dynamically adjust its compression strategy based on context, lacking content adaptability.

\textbf{Hardware-Aligned Sparsity.}
Work such as Native Sparse Attention~\citep{yuan2025nativesparseattentionhardwarealigned} designs regularized sparse patterns for modern accelerators, achieving high computational efficiency through hardware-friendly block-sparse structures. However, the core deficiency of such methods lies in their static nature. Fixed sparse patterns cannot adapt to the dynamic changes in input content, leading to potential misallocation of computational resources to non-critical regions while neglecting genuinely important information.

\subsection{Content-Aware Selective Computation}

As model capabilities have advanced, research has shifted towards content-aware selective computation, enabling models to learn autonomously where to focus their attention.

\textbf{KV Eviction.}
Methods like H2O~\citep{zhang2023h2o} and SnapKV~\citep{li2024snapkv} save memory and computation by evicting "unimportant" tokens from the KV cache. These approaches typically rely on heuristics such as attention scores or access frequency to decide which tokens to retain. While effective in some scenarios, these heuristics can lead to erroneous eviction decisions, permanently losing critical information, especially in complex reasoning tasks that require long-distance backtracking.

\textbf{Token Selection.}
Another class of methods actively selects a small subset of tokens for attention computation through techniques like sampling~\citep{chen2024magicpig}, hashing~\citep{desai2024hashattention}, or clustering~\citep{liu2024clusterkv}. While theoretically appealing, these methods face two major challenges in practice. First, discrete selection operations (such as sampling and hashing) are often non-differentiable, which impedes end-to-end training and prevents the model from learning optimal sparse patterns. Second, token-granular selection strategies disrupt memory access continuity, rendering them incompatible with modern efficient attention implementations like FlashAttention, leading to low hardware utilization and a decrease in both training and inference speed.

In summary, existing sparse attention mechanisms face an inherent trade-off between efficiency and effectiveness. Position-based methods, while efficient, lack flexibility and content awareness. Content-based methods, while more intelligent, are often limited by non-differentiable operations and inefficient hardware implementations. This dilemma highlights the urgent need for a new attention paradigm that can leverage structured sparsity for efficient computation while enabling content-aware selection through a trainable, dynamic mechanism.

\section{Background}
\label{sec:background}

\begin{figure}[!tbp]
    \centering
    \includegraphics[width=\linewidth]{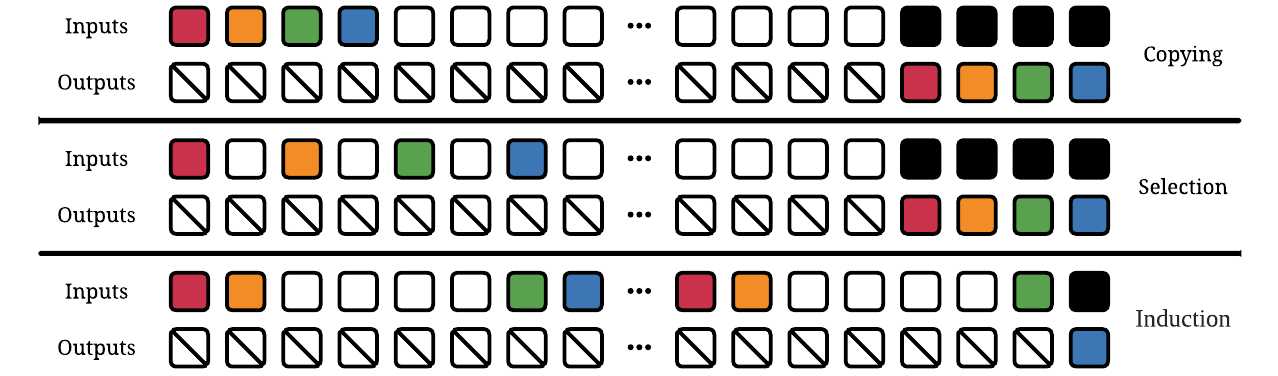}
    \caption{
    \textbf{Sparsity in Language Modeling Tasks}.
    The tasks of Copy, Select, and Induce are three essential tasks for language modeling. The Copy task requires maintaining a fixed distance between input elements and output elements, the Select task involves selectively remembering or ignoring certain elements based on the input, and the Induce task requires retrieving answers through associative recall based on context. Where the colored parts represent the tokens that the model needs to remember in the current time step of inference, the black parts represent the output tokens that the model needs to predict based on the input, and the white parts represent irrelevant tokens that can be filtered out.
    }
    \label{fig:content_aware}
\end{figure}

\subsection{Language Modeling Tasks}

\textbf{Sparsity in Language Modeling.}
As illustrated in Figure~\ref{fig:content_aware}, long-context language modeling can be decomposed into three fundamental tasks: Copying~\citep{romero2021ckconv}, Selecting~\citep{arjovsky2016unitary}, and Inducing~\citep{olsson2022context}. The Copy task requires preserving fixed-distance relationships between input and output tokens. The Select task involves selectively retaining or discarding information based on its content. The Induce task necessitates retrieving information via associative recall from the context. Each of these tasks is characterized by a distinct sparsity pattern: the Copy task exhibits \textit{positional sparsity}, attending only to tokens at fixed distances; the Select task demonstrates \textit{content sparsity}, focusing on tokens with specific content; and the Induce task relies on \textit{associative sparsity}, where attention is directed only to key-value pairs relevant to the query. These inherent sparsity patterns provide a strong theoretical foundation for designing more efficient attention mechanisms.

\subsection{Multi-Head Attention}

\textbf{QKV Projection.}
In the Transformer architecture~\citep{vaswani2017attention}, the input is first transformed into $Q$, $K$, $V$. For the hidden state of the $t$-th token in a sequence of length $n$, denoted as $h_t \in \mathbb{R}^{d_{model}}$, the linear projections are performed using weight matrices $W^{Q}$, $W^{K}$, and $W^{V}$ to obtain $q_t$, $k_t$, and $v_t$, respectively, as shown in Equation~\ref{eq:qkv_proj}. These projections map the input representation into distinct subspaces for each of the $n_h$ attention heads, allowing each head to focus on different aspects of the input. The weight matrices shape the projections to have a dimension of $d_h$ per head.

\vspace{-0.75em}
\begin{equation}
\begin{aligned}
q_{t} &= h_t W^{Q} \quad
where \quad h_t \in \mathbb{R}^{d_{model}} \quad W^{Q} \in \mathbb{R}^{d_{model} \times n_{h} \times d_{h}} \quad q_{t} \in \mathbb{R}^{n_{h} \times d_{h}} \\
k_{t} &= h_t W^{K} \quad
where \quad h_t \in \mathbb{R}^{d_{model}} \quad W^{K} \in \mathbb{R}^{d_{model} \times n_{h} \times d_{h}} \quad k_{t} \in \mathbb{R}^{n_{h} \times d_{h}} \\
v_{t} &= h_t W^{V} \quad
where \quad h_t \in \mathbb{R}^{d_{model}} \quad W^{V} \in \mathbb{R}^{d_{model} \times n_{h} \times d_{h}} \quad v_{t} \in \mathbb{R}^{n_{h} \times d_{h}}
\end{aligned}
\label{eq:qkv_proj}
\end{equation}
\vspace{-0.75em}

\textbf{Key-Value Concatenation.}
During autoregressive generation, the key-value pairs of historical tokens are cached to prevent redundant computations. As shown in Equation~\ref{eq:kv_concat}, the cached key and value matrices from past tokens are concatenated with the key-value representations of the current token to form the complete key matrix $K$ and value matrix $V$. By maintaining and updating this cache, a complete context window spanning all tokens from position 1 to the current position $t$ is constructed, enabling the model to access and utilize the full sequence history.

\vspace{-0.75em}
\begin{equation}
\begin{aligned}
k &= \operatorname{concat}([k_{1}, \ldots, k_{t}]) \quad
where \quad k \in \mathbb{R}^{n_h \times n \times d_h} \\
v &= \operatorname{concat}([v_{1}, \ldots, v_{t}]) \quad
where \quad v \in \mathbb{R}^{n_h \times n \times d_h}
\label{eq:kv_concat}
\end{aligned}
\end{equation}

\section{Method}
\label{sec:methods}

\begin{figure}[t]
    \centering
    \includegraphics[width=\linewidth]{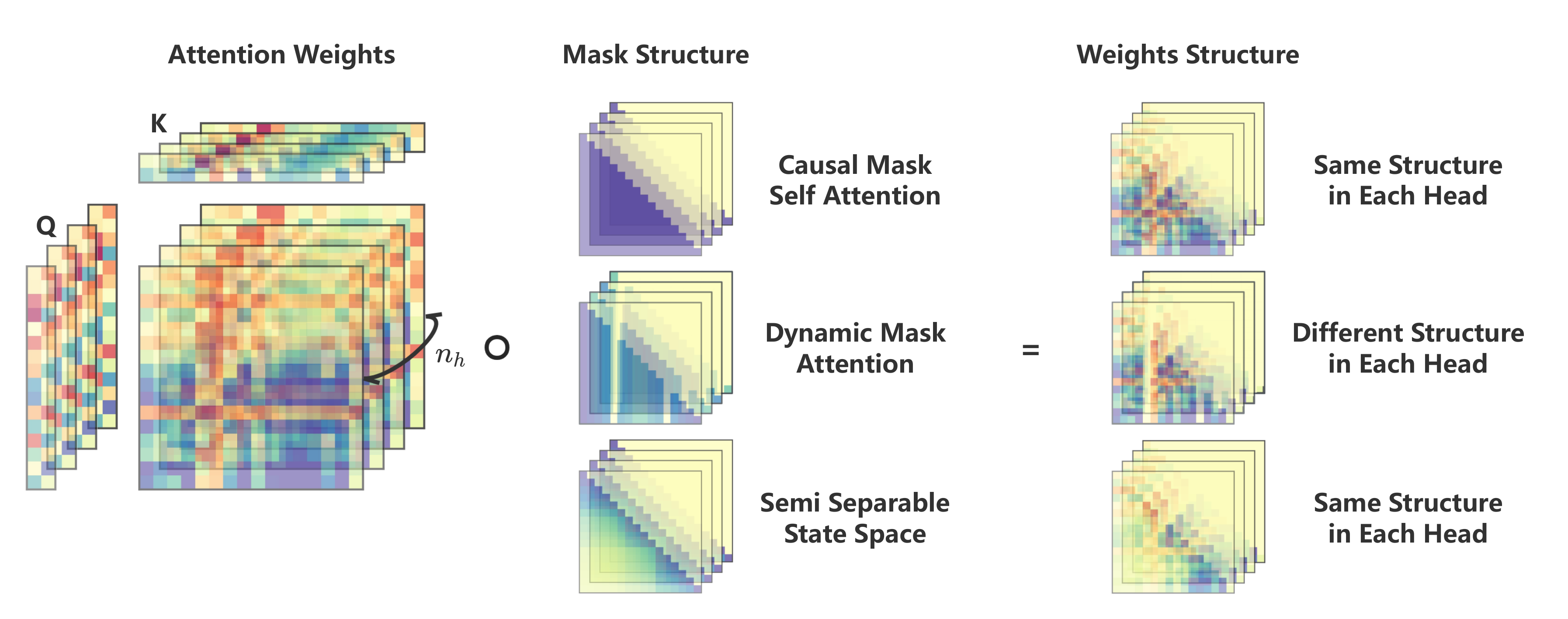}
    \caption{
    \textbf{Dynamic Mask Attention Structure}.
    It demonstrates the mask structure and weight structure of Dynamic Mask Attention in the multi-head case. Unlike the same and redundant mask and weight structures in Self-Attention and State-Space, the mask structure of DMA is dynamically adjusted through content awareness, where each head's mask can be different. This allows DMA to achieve different attention weight distributions in each head, enabling the model to maximize the utilization of each subspace in multi-head attention and focus on different tokens in each head.
    }
    \label{fig:dma}
\end{figure}

As discussed in Section~\ref{sec:related_work}, existing attention mechanisms are confronted with numerous challenges when processing long sequences, including high computational complexity, substantial memory requirements, the rigidity of static masks, and the presence of non-differentiable components that impede end-to-end training. To address these issues, we introduce Dynamic Mask Attention, a mechanism that strategically leverages the inherent sparsity patterns of language modeling. As illustrated in Figure~\ref{fig:dynamic_mask_attn}, DMA is composed of two core components: content-aware dynamic sparse masking and position-aware sparse attention weight computation. The former utilizes value representations to dynamically generate masks that determine which historical tokens each attention head should attend to, while the latter performs efficient sparse attention computations guided by these masks. This dual-component design allows DMA to maintain focus on critical information while adapting to varying contextual dependencies. Furthermore, as shown in Figure~\ref{fig:dma}, DMA generates a unique mask structure for each attention head, enabling the model to capture diverse content patterns across different representational subspaces. A sample PyTorch implementation is provided in Listing~\ref{lst:dynamic_mask_attn_code} for reference.

\subsection{Generate Content-Aware Dynamic Mask}

The content-aware dynamic mask constitutes the central innovation of DMA. It operates by analyzing the content features embedded within the value representations to determine which historical information is relevant to the current query. For each attention head at the current time step, this mechanism generates a unique dynamic mask that directs the subsequent attention weight computation to focus exclusively on the most critical key positions.

To sample the value vector representations, we introduce a sampling weight matrix $\Delta \in \mathbb{R}^{n_{h} \times d_{h} \times n_{h}}$, a per-head gating coefficient $A \in \mathbb{R}^{n_{h}}$, and a non-negative activation function $\tau(\cdot)$. As detailed in Equation~\eqref{eq:delta_calc}, the process begins with a tensor contraction, $v \cdot \Delta$, which projects each token's $d_h$-dimensional value vector into a scalar representation, serving as an initial estimate of its importance. Subsequently, the activation function $\tau(\cdot)$ ensures these scores are non-negative, preventing signal cancellation. The gating coefficient $A$ then scales the importance scores for each head, allowing the model to learn varying degrees of sparsity across heads, where $A$ can be made query-dependent, enabling the sampling scores to adapt based on the input. Finally, an exponential function amplifies the differences between scores and maps them to a positive value space, which facilitates the learning of gating effects and yields the final scores, $\delta \in \mathbb{R}^{n_{h} \times n}$.

\begin{equation}
\begin{aligned}
\delta &= \exp(\tau(v \cdot \Delta) \times A) \quad
where \quad \Delta \in \mathbb{R}^{n_h \times d_h \times n_h} \quad A \in \mathbb{R}^{n_h} \quad \delta \in \mathbb{R}^{n_h \times n} \\
\end{aligned}
\label{eq:delta_calc}
\end{equation}

Subsequently, as defined in Equation~\eqref{eq:mask_combine}, a sparsification function $f(\cdot)$ is applied. This function identifies whether each score $\delta_{n_h,j}$ ranks within the $top_w$ for its respective head. Scores within the $top_w$ are retained to preserve the gradient flow, while all other scores are set to $-\infty$, effectively nullifying their contribution in the subsequent softmax operation. For causal language modeling, this function can incorporate a causal mask via broadcasting to avoid additional memory overhead. This process yields the final dynamic mask, $m_t \in \mathbb{R}^{n_{h} \times n}$.

\begin{equation}
\begin{aligned}
m_t &= f(\delta) \quad where \quad m_t \in \mathbb{R}^{n_h \times n} \\
&= \begin{bmatrix}
    f(\sum_{j=1}^{n} \delta_{1,j}) \\
    f(\sum_{j=1}^{n} \delta_{2,j}) \\
    \vdots \\
    f(\sum_{j=1}^{n} \delta_{n_h,j}) \\
\end{bmatrix} \quad
where \quad f(\delta_{n_h,j}) = \begin{cases}
    \delta_{n_h,j} & \text{if } \delta_{n_h,j} \in top_w(\delta_{n_h}) \\
    -\infty & \text{otherwise}
\end{cases}
\label{eq:mask_combine}
\end{aligned}
\end{equation}

This approach offers three significant advantages. First, by sampling importance scores from value representations, the model can more accurately focus on semantically critical tokens, regardless of their distance. This mitigates the risk of overlooking important long-range dependencies, a common issue with methods relying solely on positional patterns. Second, the combination of the gating coefficient $A$ and independent $top_w$ selection allows different attention heads to specialize in distinct functions, such as local, long-range, and global, thereby improving the breadth of representational coverage. Third, the sparse selection mechanism is inherently effective during training, eliminating the need for post-hoc pruning and preserving the model's learned retrieval capabilities.
The kernel implementation, illustrated in Figure~\ref{fig:dynamic_mask_attn} (left), is designed for efficiency. In an outer loop, the sampling weight $\Delta$ and gating coefficient $A$ are loaded into high-speed SRAM. In an inner loop, a zero-order hold method iteratively processes blocks of the value matrix $V$ from SRAM to generate content-aware masks for the key matrix $K$. These masks are then causally broadcast to match the length of the query matrix $Q$ in HBM, avoiding memory usage with quadratic complexity. Finally, the mask blocks are concatenated to form the complete content-aware dynamic mask.

\subsection{Compute Position-Aware Sparse Weights}

The second core component of DMA is the position-aware sparse weight computation. As outlined in Algorithm~\ref{alg:fdma_fwd}, this component leverages the previously generated dynamic mask to sparsify the computation of scaled dot-product attention weights, effectively reducing the per-step complexity from $O(n d_h)$ to $O(w d_h)$.

For the query at step $t$, $q_t \in \mathbb{R}^{n_h \times d_h}$, and the complete key-value pairs, $K, V \in \mathbb{R}^{n_h \times n \times d_h}$, the entire computation flow is detailed in Equation~\eqref{eq:attention_output}. Initially, for each attention head $n_h$, the scaled dot-product between the query and keys, $q_{t} K^{\top}$, is computed and then element-wise multiplied by the previously constructed dynamic mask $m_t$. The scaling factor $\sqrt{d_h}$ is crucial here as it prevents the dot products from becoming excessively large, which could push the softmax function into a saturated region with minimal gradients. After applying the mask, the softmax function normalizes the results to produce attention weights $p_{n_h,j}$. Notably, when a mask value $m_{n_h,j} = -\infty$, the corresponding attention weight $p_{n_h,j} \approx 0$, effectively skipping computations for masked positions and filling them with zeros. This ensures that the model focuses solely on relevant unmasked contexts. The attention weights for each head are then multiplied by the value vectors and summed to produce the final context vector $o_{t} \in \mathbb{R}^{n_h \times d_h}$, where each row captures different contextual patterns and dependencies. The multi-head mechanism, combined with dynamic masking, allows the model to attend to various patterns in parallel across the sequence. This output integrates information from all attention heads, forming a rich hierarchical context representation that effectively captures dependencies at varying distances within the sequence history. It is important to note that this method can approximate full attention when $n_h \times w \leq n$, while maintaining computational efficiency.

\begin{equation}
\begin{aligned}
o_{t} &= \operatorname{softmax}(\frac{q_{t} k^{\top}} {\sqrt{d_{h}}} \circ m_t) v \quad where \quad p_{t} \in \mathbb{R}^{n_h \times n} \quad o_{t} \in \mathbb{R}^{n_h \times d_h} \\
&= \begin{bmatrix}
    \sum_{j=1}^{n} p_{1,j} \cdot v_{1,j} \\
    \sum_{j=1}^{n} p_{2,j} \cdot v_{2,j} \\
    \vdots \\
    \sum_{j=1}^{n} p_{n_h,j} \cdot v_{n_h,j} \\
\end{bmatrix} \quad 
where \quad p_{n_h,j} = \begin{cases}
    \frac{
        \exp(\frac{q_{n_h} \cdot k_{n_h,j}^{\top}}{\sqrt{d_h}} + m_{n_h,j})
    }{
        \sum_{j'=1}^{n} \exp(\frac{q_{n_h} \cdot k_{n_h,j'}^{\top}}{\sqrt{d_h}} + m_{n_h,j'})
    } & \text{if } m_{n_h,j} \neq -\infty \\
    0 & \text{if } m_{n_h,j} = -\infty
\end{cases}  \\
\label{eq:attention_output}
\end{aligned}
\end{equation}

\begin{algorithm}[!ht]
  \small
  \caption{Flash Dynamic Mask Attention Forward Pass Per Head}
  \label{alg:fdma_fwd}
  \begin{algorithmic}
    \REQUIRE Matrices $\vQ, \vK, \vV \in \mathbb{R}^{N \times d_h}, \vM \in \mathbb{R}^{N}$ in HBM. Set block sizes $B$.
    \STATE Initialize $\vO = (0)_{N \times d_h} \in \mathbb{R}^{N \times d_h}, \ell = (0)_N \in \mathbb{R}^{N}, m = (-\infty)_N \in \mathbb{R}^{N}$ in HBM.
    \STATE Divide $\vQ$ into $T_r$ blocks of size $B \times d_h$ each, and divide $\vK, \vV$ into $T_c$ blocks of size $B \times d_h$ each.
    \STATE Divide $\vM$ into $T_c$ blocks of size $B$ each.
    \STATE Divide $\vO$ into $T_r$ blocks of size $B \times d_h$ each, divide $\ell$ into $T_r$ blocks of size $B$ each, divide $m$ into $T_r$ blocks of size $B$ each.
    \FOR{$1 \le j \le T_c$} 
      \STATE Load $\vM_j$ from HBM to SRAM.
      \STATE Compute $\mathrm{active}_j = \mathrm{Judge}(\vM_j)$.
      \IF{$\mathrm{active}_j = 0$}
        \STATE Advance stream pointers for $\vK_j, \vV_j$, \textbf{continue}.
      \ENDIF
      \STATE Load $\vK_j, \vV_j$ from HBM to SRAM.
      \FOR{$1 \le i \le T_r$}
        \STATE Load $\vQ_i, \vO_i, \ell_i, m_i$ from HBM to SRAM.
        \STATE Compute $\vS_{ij} = \vQ_i \vK_j^T \times d_h^{-0.5} + \vM_j \in \mathbb{R}^{B \times B}$.
        \STATE Compute $\tilde{m}_{ij} = \mathrm{rowmax}(\vS_{ij}) \in \mathbb{R}^{B}$, $\tilde{\vP}_{ij} = \exp(\vS_{ij} - \tilde{m}_{ij}) \in \mathbb{R}^{B \times B}$, $\tilde{\ell}_{ij} = \mathrm{rowsum}(\tilde{\vP}_{ij}) \in \mathbb{R}^{B}$.
        \STATE Compute $m_i^{\mathrm{new}} = \max(m_i, \tilde{m}_{ij}) \in \mathbb{R}^{B}$, $\ell_i^{\mathrm{new}} = e^{m_i - m_i^{\mathrm{new}}} \ell_i + e^{\tilde{m}_{ij} - m_i^{\mathrm{new}}} \tilde{\ell}_{ij} \in \mathbb{R}^{B}$.
        \STATE Write $\vO_i \leftarrow \diag(\ell_i^{\mathrm{new}})^{-1}(\diag(\ell_i) e^{m_i - m_i^{\mathrm{new}}} \vO_i + e^{\tilde{m}_{ij} - m_i^{\mathrm{new}}}\tilde{\vP}_{ij} \vV_j)$ to HBM.
        \STATE Write $\ell_i \leftarrow \ell_i^{\mathrm{new}}$, $m_i \leftarrow m_i^{\mathrm{new}}$ to HBM.
      \ENDFOR
    \ENDFOR
    \STATE Return $\vO, \ell, m$.
  \end{algorithmic}
\end{algorithm}

This method offers three key advantages. First, the mask prunes the set of candidate tokens before the matrix multiplication and softmax operations. This avoids the inefficiency of pseudo-sparsity, where computations are performed for all tokens only to be zeroed out afterward. Second, unlike methods that perform key-value selection by discarding tokens, our approach preserves the full sequence. This ensures that the complete global context remains available for all attention heads to access as needed. Third, the kernel implementation can perform block-level skipping by loading a block of the mask to check if all positions within it are masked. If so, the entire block is skipped, avoiding unnecessary memory loads and matrix multiplication operations.
The kernel implementation, depicted in Figure~\ref{fig:dynamic_mask_attn} (right), is optimized for this process. In an outer loop, blocks of the $K$ and $V$ matrices are loaded into SRAM. In an inner loop, blocks of the $Q$ matrix are loaded, and if the corresponding $K$ block is not entirely masked, the attention weights are computed and the output is written back to HBM. If a position in the $K$ block is masked, its attention weight is set to zero, and the computation for that position is skipped, resulting in position-aware sparse attention weights.

\subsection{Fully Gradient Flow}

\begin{algorithm}[!ht]
  \small
  \caption{Flash Dynamic Mask Attention Backward Pass Per Head}
  \label{alg:fdma_bwd}
  \begin{algorithmic}
    \REQUIRE Matrices $\vQ, \vK, \vV, \vO, \vdO \in \mathbb{R}^{N \times d_h}, \vM, \vdM \in \mathbb{R}^{N}$ in HBM, vectors $\ell, m \in \mathbb{R}^N$ in HBM. Set block sizes $B$.
    \STATE Divide $\vQ$ into $T_r$ blocks of size $B \times d_h$ each, and divide $\vK, \vV$ in to $T_c$ blocks of size $B \times d_h$ each.
    \STATE Divide $\vM$ into $T_c$ blocks of size $B$ each.
    \STATE Divide $\vO, \vdO$ into $T_r$ blocks of size
    $B \times d_h$ each, divide $\ell$ into $T_r$ blocks of size
    $B$ each, and divide $m$ into $T_r$ blocks of size $B$ each.
    \STATE Initialize $\vdQ = (0)_{N \times d_h}$ in HBM and divide it into $T_r$ blocks of size $B \times d_h$ each.
    \STATE Initialize $\vdK = (0)_{N \times d_h}, \vdV = (0)_{N \times d_h}$ in HBM and divide $\vdK, \vdV$ in to $T_c$ blocks of size $B \times d_h$ each.
    \STATE Initialize $\vdM$ = (0)$_{N}$ in HBM and divide it into $T_c$ blocks of size $B$ each.
    \FOR{$1 \le j \le T_c$}
      \STATE Load $\vM_j$ from HBM to SRAM.
        \STATE Compute $\mathrm{active}_j = \mathrm{Judge}(\vM_j)$.
        \IF{$\mathrm{active}_j = 0$}
          \STATE Advance stream pointers for $\vK_j, \vV_j$, \textbf{continue}.
        \ENDIF
      \STATE Load $\vK_j, \vV_j$ from HBM to SRAM.
      \STATE Initialize $\tilde{\vdK}_j = (0)_{B \times d_h}, \tilde{\vdV}_j = (0)_{B \times d_h}, \tilde{\vdM}_j = (0)_{B}$ on SRAM.
      \FOR{$1 \le i \le T_r$}
        \STATE Load $\vQ_i, \vO_i, \vdO_i, \vdQ_i, \ell_i, m_i$ from HBM to SRAM.
        \STATE Compute $\vS_{ij} = \vQ_i \vK_j^T \times d_h^{-0.5} + \vM_j \in \mathbb{R}^{B \times B}$.
        \STATE Compute $\vP_{ij} = \diag(\ell_i)^{-1}\exp(\vS_{ij} - m_{i}) \in \mathbb{R}^{B \times B}$.
        \STATE Compute
        $\tilde{\vdV_j} \leftarrow \tilde{\vdV_j} + \vP_{ij}^\top \vdO_i \in \mathbb{R}^{B \times d_h}$.
        \STATE Compute
        $\vdP_{ij} = \vdO_{i} \vV_j^\top \in \mathbb{R}^{B \times B}$.
        \STATE Compute $D_{i} = \mathrm{rowsum}(\vdO_i \circ \vO_i) \in \mathbb{R}^{B}$.
        \STATE Compute $\vdS_{ij} = \vP_{ij} \circ (\vdP_{ij} - D_i) \in \mathbb{R}^{B \times B}$.
        \STATE Compute $\tilde{\vdM}_{j} \leftarrow \tilde{\vdM}_j + \mathrm{rowsum}(\vdS_{ij}) \in \mathbb{R}^{B}$.
        \STATE Write $\vdQ_{i} \leftarrow \vdQ_i + \vdS_{ij} \vK_j \times d_h^{-0.5} \in \mathbb{R}^{B \times d_h}$ to HBM.
        \STATE Compute $\tilde{\vdK}_j \leftarrow \tilde{\vdK}_j + \vdS_{ij}^\top \vQ_i \times d_h^{-0.5} \in \mathbb{R}^{B \times d_h}$.
      \ENDFOR
      \STATE Write $\vdK_j \leftarrow \tilde{\vdK_j}, \vdV_j \leftarrow \tilde{\vdV_j}, \vdM_j \leftarrow \vdM_j + \tilde{\vdM}_j$ to HBM.
    \ENDFOR
    \STATE Return $\vdQ, \vdK, \vdV, \vdM$.
  \end{algorithmic}
\end{algorithm}

Finally, the entire backward process is outlined in Algorithm~\ref{alg:fdma_bwd}. We ensure that the introduced dynamic mask and sparse weights do not block gradients, and the gradients of the retained attention paths are strictly consistent with those of full attention. They can flow completely to all inputs and parameters without gradient discontinuity issues caused by discrete operations, supporting end-to-end training while minimizing redundant costs.

For clarity, our derivation considers a single attention head $h$ at a single time step $t$. Let $\mathcal{I}_h \subset \{1, \dots, n\}$ be the set of $w$ indices selected for this head. For unselected indices $j \notin \mathcal{I}_h$, the mask value is treated as $m_{h,j}=-\infty$. The key intermediate variables in the forward pass are defined in Equation~\eqref{eq:forward_pass}.

\vspace{-1em}
\begin{equation}
\begin{aligned}
s_{h,j} &= \frac{q_h \cdot k_{h,j}}{\sqrt{d_h}} + m_{h,j} \quad
p_{h,j} &= 
\begin{cases}
\dfrac{\exp(s_{h,j})}{\sum_{j'\in \mathcal{I}_h} \exp(s_{h,j'})} & j \in \mathcal{I}_h\\
0 & j \notin \mathcal{I}_h
\end{cases} \quad
o_h &= \sum_{j\in \mathcal{I}_h} p_{h,j} v_{h,j}
\end{aligned}
\label{eq:forward_pass}
\end{equation}
\vspace{-1em}

In the backward pass, let the upstream gradient of the loss function $L$ with respect to the head's output $o_h$ be $do_h = \frac{\partial L}{\partial o_h} \in \mathbb{R}^{d_h}$. As shown in Equation~\eqref{eq:grad_v}, the gradient for $v$ is computed by distributing $do_h$ to the selected vectors $v_{h,j}$ in proportion to their attention weights $p_{h,j}$, while the gradients for unselected positions are zero.

\vspace{-1em}
\begin{equation}
\frac{\partial L}{\partial v_{h,j}} = 
\begin{cases}
p_{h,j} \, do_h & j \in \mathcal{I}_h \\
0 & j \notin \mathcal{I}_h
\end{cases}
\label{eq:grad_v}
\end{equation}
\vspace{-1em}

Next, we compute the gradient for the scores $s_{h,j}$. The gradient of the attention weights $p_{h,j}$ with respect to their inputs is $dp_{h,j} = v_{h,j} \cdot do_h$. Using the standard softmax Jacobian, we can derive the gradient for $s_{h,j}$, denoted as $ds_{h,j}$, as shown in Equation~\eqref{eq:grad_s}. For masked positions where $p_{h,j}=0$, the gradient $ds_{h,j}$ is naturally zero.

\vspace{-1em}
\begin{equation}
ds_{h,j} = p_{h,j} (dp_{h,j} - \sum_{j'\in \mathcal{I}_h} p_{h,j'} \times dp_{h,j'})
\label{eq:grad_s}
\end{equation}
\vspace{-1em}

Because $s_{h,j}$ is an additive combination of $q_h \cdot k_{h,j}$ and $m_{h,j}$, the gradient is distributed directly. The gradient for the mask $m_{h,j}$ is simply $ds_{h,j}$, as shown in Equation~\eqref{eq:grad_m}. This ensures that gradients can flow directly to $\Delta$ and $A$.

\vspace{-1em}
\begin{equation}
\frac{\partial L}{\partial m_{h,j}} = ds_{h,j}
\label{eq:grad_m}
\end{equation}
\vspace{-1em}

Finally, as shown in Equation~\eqref{eq:grad_qk}, the gradients for $q_h$ and $k_{h,j}$ are obtained by backpropagating $ds_{h,j}$ through the computation path. Crucially, the gradient calculations only involve the selected index set $\mathcal{I}_h$, thereby reducing computation.

\vspace{-1em}
\begin{equation}
\begin{aligned}
\frac{\partial L}{\partial q_h} = \sum_{j\in\mathcal{I}_h} ds_{h,j} \frac{k_{h,j}}{\sqrt{d_h}},
\qquad
\frac{\partial L}{\partial k_{h,j}} = ds_{h,j} \frac{q_h}{\sqrt{d_h}}
\end{aligned}
\label{eq:grad_qk}
\end{equation}
\vspace{-1em}

Our approach has several significant advantages. First, for the selected positions, the gradients are identical to those of full attention, and DMA only prunes the operator chain for positions whose contributions can be ignored, ensuring expressiveness. Then, only second-order correlation information is propagated to $\mathcal{I}_h$, improving bandwidth utilization. The gating parameter $A$ and weight $\Delta$ directly receive attention weights as gradients, quickly shaping head specialization. Finally, the equivalence relation $dM = dS$ allows the kernel to only recompute the local $S$ without storing additional intermediate mask gradient tensors.

\section{Experiments}
\label{sec:experiments}

We will validate the efficiency and effectiveness of Dynamic Mask Attention, as detailed in Section~\ref{sec:methods}, in handling long contexts through its content-aware dynamic sparse mask and position-aware dynamic sparse weight computation.

\subsection{Experimental Settings}
\label{sec:training_settings}

\paragraph{Baselines.}
To thoroughly evaluate DMA, we benchmarked it against representative baselines surveyed in Section~\ref{sec:related_work}. First, we compared DMA with various mainstream attention variants in terms of pre-training perplexity at different model scales, further validating DMA's advantage in long-sequence information retrieval through the challenging multi-query associative recall task, and tested the efficiency of our kernel compared to other optimized implementations in Section~\ref{sec:variants_comparison}. Second, we compared DMA, NSA, and MHA on a 1.7B parameter Transformer model pre-trained and supervisedly fine-tuned on 40B tokens, evaluating them on downstream benchmark tasks and needle-in-a-haystack tests in Section~\ref{sec:performance_comparison}.

\paragraph{Training Settings.}
All experiments were conducted using the open-source PyTorch images~\citep{nv2022pytorch} and the Transformers framework~\citep{wolf-etal-2020-transformers}. For model configuration, we consistently employed the NeoX tokenizer~\citep{black2022gpt}, the AdamW optimizer~\citep{Loshchilov2017FixingWD}, and the WSD learning rate scheduler~\citep{hägele2024scalinglawscomputeoptimaltraining}, while strictly adhering to the Optimal Hyperparameter Scaling Law~\citep{li2025predictablescalei} and the Chinchilla~\citep{hoffmann2022empirical} standard protocol throughout our training on the SmolLMCorpus~\citep{benallal2024smollmcorpus} dataset. For evaluation frameworks, we utilized the LM evaluation harness~\citep{eval-harness} from EleutherAI for perplexity tasks, and the lighteval~\citep{Clémentine2023lighteval} from HuggingFace for downstream tasks.

\subsection{Variants Comparison}
\label{sec:variants_comparison}

\begin{figure}[!t]
    \centering
    \includegraphics[width=0.92\linewidth]{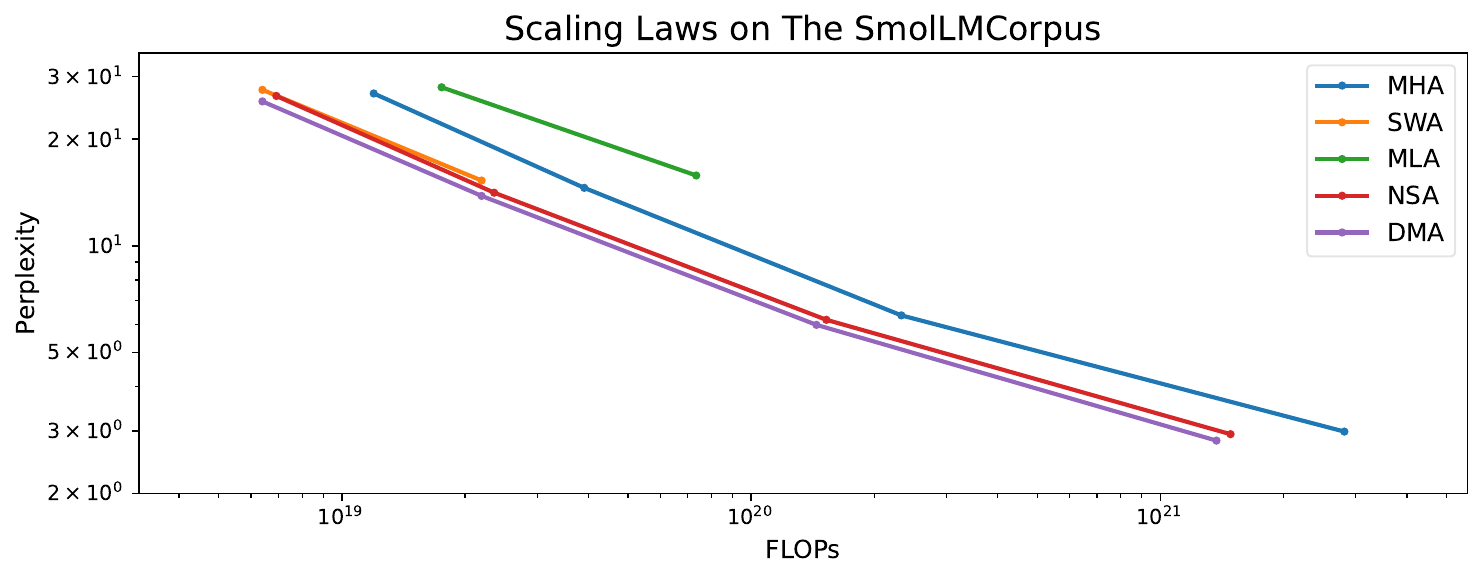}
    \caption{
    \textbf{Scaling Laws}.
    The perplexity performance of different self-attention variants on SmolLMCorpus at different parameter scales. For suboptimal variants like SWA and MLA, we omit them for clarity. Compared to other variants, our Dynamic Mask Attention has a Pareto advantage in performance.
    }
    \label{fig:scaling_laws}
\end{figure}

\paragraph{Scaling Perplexity.}
First, we present the comparison of the perplexity performance of different self-attention variants at various parameter scales in Figure~\ref{fig:scaling_laws}. This experiment includes the baseline~\footnotemark[1], sliding window attention~\footnotemark[2] driven by static mask structures, multi-head latent attention~\footnotemark[3] driven by low-rank decomposition approximations, native sparse attention~\footnotemark[4] driven by hardware content adaptation, and our proposed Dynamic Mask Attention. These experiments were conducted on the SmolLMCorpus dataset, with model sizes ranging from 80M to 1.7B parameters, and the experimental configurations are detailed in Table~\ref{tab:scaling_laws_configurations}. Our experimental results validate that Dynamic Mask Attention maintains the best performance across various scales. We speculate that this advantage primarily stems from DMA's ability to adaptively focus on key information in the input sequence, effectively avoiding the lost in middle~\citep{liu2023lostmiddlelanguagemodels} problem.

\begin{figure}[!t]
    \centering
    \includegraphics[width=\linewidth]{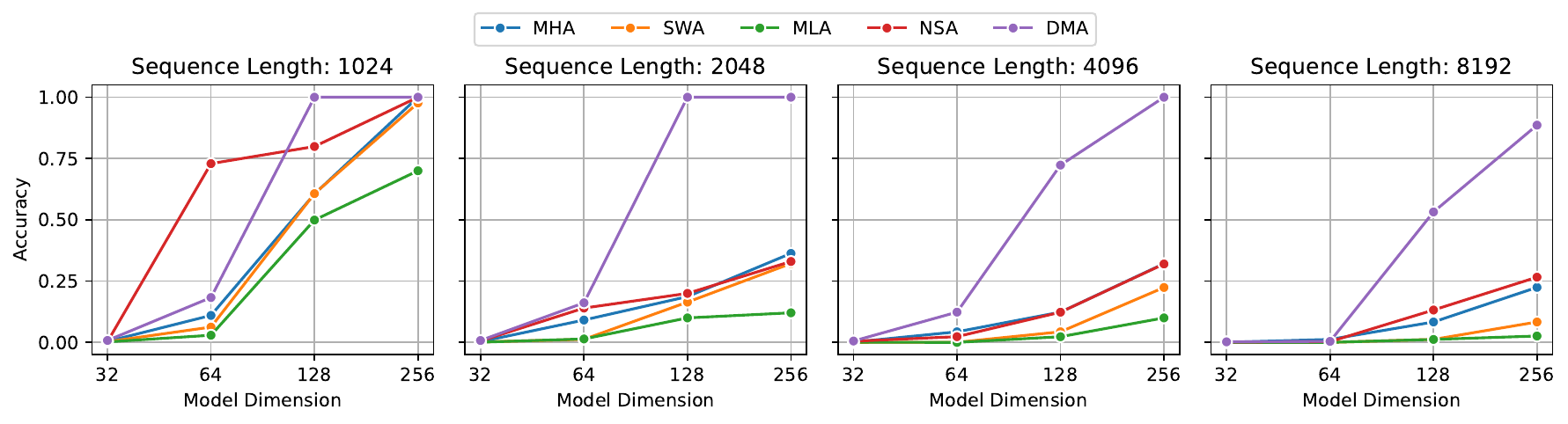}
    \caption{
    \textbf{Multi-Query Associative Recall}.
    This is a more challenging version of the original multi-query associative recall task~\citep{arora2024zoology}, which includes longer sequence lengths and smaller model dimensions. Dynamic Mask Attention maintains good performance in most cases.
    }
    \label{fig:mqar}
\end{figure}

\paragraph{Associative Recall.}
To further validate the ability of different attention variants in long-sequence information retrieval, we designed a more challenging variant of the multi-query associative recall task~\citep{arora2024zoology}, which includes longer sequence lengths and smaller model dimensions. This task assesses the ability of language models to retrieve information within their context. Specifically, it provides key-value pairs to the autoregressive model, prompting the model to generate the correct value when displaying previously seen keys. To increase the difficulty of the task, we used 512 key-value pairs in the experiment. We employ sliding window attention, native sparse attention, and dynamic mask attention, all with a window size of 512. This approach replaces non-query/key/value parts with random tokens, forcing the model to locate relevant information precisely rather than relying on contextual clues. The experimental dataset comprises 250,000 training samples and 1,000 test samples, with all models trained for 100 epochs to ensure sufficient convergence. As shown in Figure~\ref{fig:mqar}, Dynamic Mask Attention performs excellently across various sequence lengths, indicating its ability to intelligently identify and focus on tokens relevant to the current state while ignoring irrelevant tokens.

\begin{figure}[!t]
    \centering
    \includegraphics[width=\linewidth]{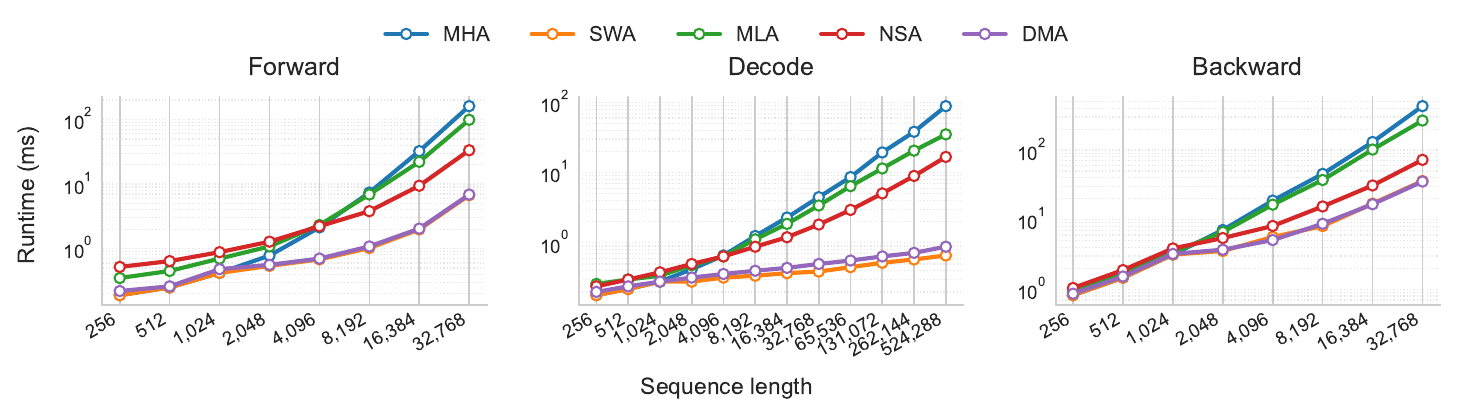}
    \caption{
    \textbf{Kernal Performance.}
    A performance comparison of efficient implementation kernels of different attention variants on an A100 GPU. Our DMA achieves significant acceleration in forward propagation, decoding, and backward propagation, maintaining the same efficiency level as SWA.
    }
    \label{fig:kernel_acceleration}
\end{figure}

\paragraph{Kernel Acceleration.}
To analyze the performance of DMA within modern efficient operator frameworks, we benchmarked the forward, decoding, and backward performance of MHA, SWA, MLA, NSA, and DMA on an A100-SXM4-80GB GPU. The results represent the average of 1,000 runs after three warm-up iterations; specific configurations and implementations can be found in Table~\ref{tab:speed_benchmark_configs}. As shown in Figure~\ref{fig:kernel_acceleration}, for the Forward pass, compared to MHA, DMA achieves speedups of approximately $26.1\times$, $10.2\times$, and $21.5\times$ at token lengths of $8192$, $16384$, and $32768$, respectively. For the Decode phase, speedups against MHA at key lengths of $65536$, $131072$, $262144$, and $524288$ are approximately $49.6\times$, $92.7\times$, and $171.1\times$, respectively. For the Backward pass, speedups against MHA at lengths of $8192$, $16384$, and $32768$ are approximately $2.5\times$, $4.4\times$, and $7.9\times$, respectively. DMA avoids large-scale redundant score and softmax backpropagation overhead by explicitly skipping masked blocks, thus providing strong acceleration capabilities across multiple critical stages.

\subsection{Performance Comparison}
\label{sec:performance_comparison}

\begin{figure}[!t]
    \centering
    \includegraphics[width=\linewidth]{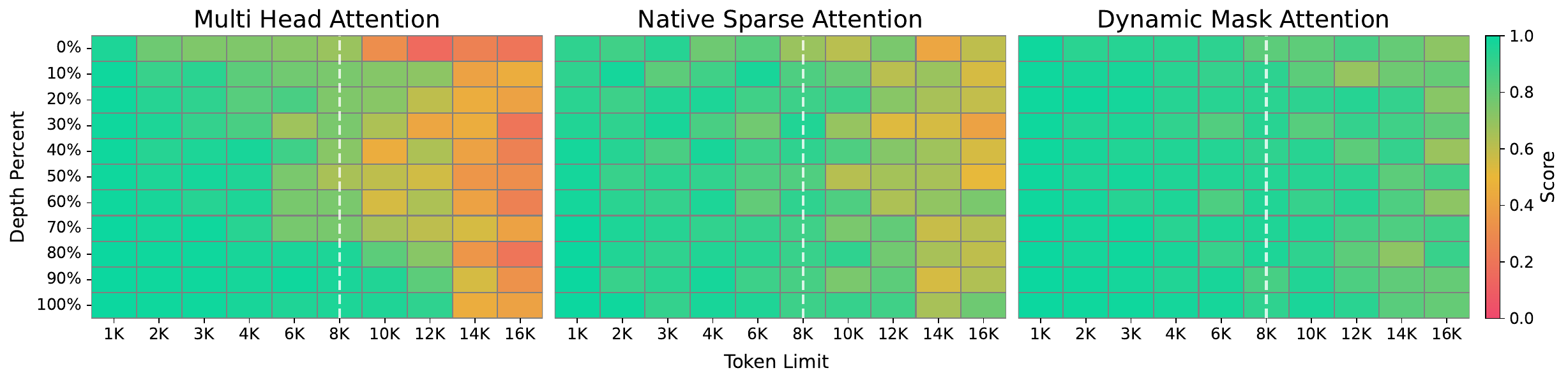}
    \caption{
    \textbf{Needle in a Haystack}.
    Comparison of needle-in-a-haystack performance between MHA, NSA, and DMA in an apples-to-apples setting. The white dotted line indicates the sequence length of the model.
    }
    \label{fig:needle_in_a_haystack}
\end{figure}

\begin{table*}[!t]
  \small
  \centering
  \caption{
    \textbf{Downstream Task Evaluations for Base Model}.
    The best results for each size are in bold, and the second-best results are unlined. DMA outperforms MHA and NSA, as well as other advanced inference sparse methods, on most tasks.
  }
  \resizebox{0.99\linewidth}{!}
  {
    \begin{tabular}{@{}ccccccccccccccc@{}}
    \toprule
    \sc{Model}  & \sc{LAMBADA}          & \sc{LAMBADA}          & \sc{MMLU}             & \sc{TriviaQA}         & \sc{ARC}              & \sc{PIQA}             & \sc{HellaSwag}        & \sc{OBQA}             & \sc{WinoGrande}       & \sc{Avg}              \\
                & \sc{ppl $\downarrow$} & \sc{acc $\uparrow$}   & \sc{acc $\uparrow$}   & \sc{acc $\uparrow$}   & \sc{acc $\uparrow$}   & \sc{acc $\uparrow$}   & \sc{acc $\uparrow$}   & \sc{acc $\uparrow$}   & \sc{acc $\uparrow$}   & \sc{acc $\uparrow$}   \\
    \midrule
    \multicolumn{10}{c}{\text{Zero-Shot}} \\
    \midrule
    MHA         & 15.22                 & 44.3                  & \underline{35.4}      & \textbf{9.4}          & \underline{53.4}      & \underline{72.9}      & 56.1                  & \textbf{37.0}         & 57.3                  & \underline{45.7}      \\
    H2O         & 15.38                 & 44.2                  & 34.8                  & 7.4                   & 53.3                  & 72.8                  & 55.6                  & 36.6                  & 56.9                  & 45.2                  \\
    InfLLM      & 15.23                 & 44.2                  & 35.1                  & 8.0                   & 53.1                  & 72.4                  & 55.8                  & 36.6                  & 56.8                  & 45.2                  \\
    Quest       & 15.43                 & 43.9                  & 35.1                  & 7.6                   & 53.1                  & 72.6                  & 56.1                  & 36.8                  & 57.2                  & 45.3                  \\
    DAM         & 15.89                 & 44.5                  & 34.6                  & 8.9                   & 52.1                  & 72.3                  & 56.2                  & 36.3                  & 56.0                  & 45.1                  \\
    Exact-Top   & 15.23                 & 44.4                  & 35.3                  & \underline{9.2}       & 53.3                  & 72.8                  & 56.0                  & \underline{36.8}      & 57.0                  & 45.6                  \\
    NSA         & \underline{14.91}     & \underline{45.2}      & 33.8                  & 8.7                   & 53.1                  & 72.8                  & \textbf{56.7}         & 36.3                  & \underline{57.8}      & 45.5                  \\
    DMA (ours)  & \textbf{14.42}        & \textbf{45.9}         & \textbf{37.0}         & 9.1                   & \textbf{55.6}         & \textbf{73.4}         & \underline{56.4}      & 36.5                  & \textbf{58.4}         & \textbf{46.5}         \\
    \midrule
    \multicolumn{10}{c}{\text{Five-Shot}} \\
    \midrule
    MHA         & 19.40                 & 40.4                  & \underline{36.8}      & \underline{13.2}      & \textbf{56.8}         & 73.2                  & 56.8                  & 38.0                  & \underline{58.6}      & 46.7                  \\
    H2O         & 19.14                 & 38.9                  & 35.7                  & 10.2                  & 56.6                  & 73.2                  & 56.4                  & 37.8                  & 58.1                  & 45.8                  \\
    InfLLM      & 19.13                 & \textbf{41.3}         & 35.9                  & 11.7                  & \underline{56.7}      & 73.3                  & 56.1                  & 38.0                  & 57.7                  & 46.3                  \\
    Quest       & 19.22                 & 40.9                  & 36.1                  & 10.9                  & 56.2                  & 73.2                  & 55.8                  & 37.9                  & 58.2                  & 46.1                  \\
    DAM         & 19.47                 & \underline{41.2}      & 35.2                  & \textbf{13.3}         & 55.1                  & 71.0                  & 54.4                  & 38.0                  & 57.2                  & 45.6                  \\
    Exact-Top   & \underline{18.22}     & 39.7                  & 36.4                  & 13.1                  & 56.3                  & 73.4                  & 56.5                  & 38.2                  & 58.5                  & 46.5                  \\
    NSA         & 21.37                 & 39.6                  & 34.6                  & 12.5                  & 56.1                  & \underline{76.0}      & \textbf{58.9}         & \underline{39.2}      & 58.3                  & \underline{46.9}      \\
    DMA (ours)  & \textbf{17.88}        & 40.9                  & \textbf{38.2}         & 12.6                  & 56.4                  & \textbf{76.6}         & \underline{58.7}      & \textbf{39.6}         & \textbf{60.4}         & \textbf{47.9}         \\
    \bottomrule
    \end{tabular}
  }
  \label{table:downstream_base}
\end{table*}

\begin{table*}[!t]
  \small
  \centering
  \caption{
    \textbf{Downstream Evaluations for Finetuned Model}.
    The best results are in bold, and the second-best results are underlined. For models supervisedly fine-tuned at a 16K sequence length, DMA outperforms other methods in most tasks.
  }
  \resizebox{\linewidth}{!}{
    \begin{tabular}{@{}ccccccccccc@{}}
    \toprule
    \sc{Method} & \sc{MMLU}             & \sc{BBH}              & \sc{GSM8K}            & \sc{MATH}             & \sc{MBPP}             & \sc{ARC}              & \sc{PIQA}             & \sc{LongBench}        & \sc{RULER}            & \sc{Avg}              \\
                & \sc{acc $\uparrow$}   & \sc{acc $\uparrow$}   & \sc{acc $\uparrow$}   & \sc{acc $\uparrow$}   & \sc{acc $\uparrow$}   & \sc{acc $\uparrow$}   & \sc{acc $\uparrow$}   & \sc{acc $\uparrow$}   & \sc{acc $\uparrow$}   & \sc{acc $\uparrow$}   \\
    \midrule
    MHA         & \textbf{46.4}         & 37.7                  & \underline{46.3}      & \textbf{11.7}         & 40.0                  & \underline{59.8}      & \underline{76.2}      & 30.2                  & \textbf{60.6}         & \underline{39.0}      \\
    H2O         & 42.4                  & 34.5                  & 44.8                  & 10.5                  & 38.5                  & 57.6                  & 74.1                  & 28.5                  & 47.8                  & 35.3                  \\
    InfLLM      & 44.2                  & 36.0                  & 45.5                  & 11.0                  & 39.2                  & 58.9                  & 75.6                  & 29.1                  & 55.3                  & 37.2                  \\
    Quest       & 44.0                  & 34.8                  & 45.0                  & 10.8                  & 38.8                  & 56.2                  & 73.7                  & 28.9                  & 50.7                  & 36.1                  \\
    DAM         & 45.0                  & 36.5                  & 46.0                  & 11.5                  & 37.2                  & 56.9                  & 72.4                  & 26.4                  & 49.2                  & 36.0                  \\
    Exact-Top   & 45.1                  & 37.2                  & 45.1                  & 11.5                  & 38.4                  & 57.3                  & 75.4                  & 28.3                  & 44.8                  & 35.7                  \\
    NSA         & 43.8                  & \textbf{38.4}         & 46.2                  & \underline{11.6}      & \underline{40.5}      & 59.1                  & 75.7                  & \underline{30.2}      & 59.6                  & 38.6                  \\
    DMA (ours)  & \underline{46.2}      & \underline{38.2}      & \textbf{46.8}         & \underline{11.6}      & \textbf{40.6}         & \textbf{59.6}         & \textbf{76.6}         & \textbf{30.7}         & \underline{60.5}      & \textbf{39.2}         \\
    \bottomrule
    \end{tabular}
  }
  \label{table:downstream_finetune}
\end{table*}

\paragraph{Downstream Benchmark Evaluations for Base Model.}
We used the Qwen3 1.7B~\citep{qwen32025} model structure as a baseline, making only modifications to the self-attention part for comparison. We first pre-trained the model on a high-quality dataset covering four domains: Web, TextBook, Code, and Math, with a total of 32 billion tokens and a sequence length of 2,048, thereby providing the model with basic language skills and general knowledge. Subsequently, we carefully selected 8B tokens packaged into sequences of length 8K. We conducted a second phase of pre-training by adjusting the RoPE base frequency from 10K to 100K~\citep{xiong2023effectivelongcontextscalingfoundation}, ensuring that the model could effectively handle longer inputs. Ultimately, we obtained three models: MHA, NSA, and DMA, and evaluated them on the following tasks: LLAMBADA~\citep{paperno2016lambada}, MMLU~\citep{hendrycks2021measuring}, TriviaQA~\citep{joshi2017triviaqa}, ARC~\citep{clark2018think}, PIQA~\citep{bisk2020piqa}, HellaSwag~\citep{zellers2019hellaswag}, OBQA~\citep{mihaylov2018can}, Winogrande~\citep{sakaguchi2021winogrande}, and the English tasks of LongBench~\citep{bai2023longbench}. We also compared several advanced inference sparse methods, including H2O~\citep{zhang2023h2o}, infLLM~\citep{xiao2024infllm}, Quest~\citep{tang2024quest}, DAM~\citep{zhang2025damdynamicattentionmask}, and Exact-Top, which first computes full attention scores using MHA and then performs sparsification based on that. The results are shown in Table~\ref{table:downstream_base}. In both zero-shot and five-shot settings, DMA outperforms the baseline on most tasks, achieving excellent overall performance.

\paragraph{Downstream Benchmark Evaluations for Finetuned Model.}
We further fine-tuned all models at a 16K sequence length with an adjusted RoPE base frequency to 400K, to further enhance the models' long-context generalization capabilities. Ultimately, we obtained three fine-tuned models: MHA, NSA, and DMA, and evaluated them on the following tasks: MMLU~\citep{hendrycks2021measuring}, BBH~\citep{suzgun2023challenging}, GSM8K~\citep{cobbe2021training}, MATH~\citep{hendrycks2020measuring}, MBPP~\citep{austin2021program}, LongBench~\citep{bai2023longbench}, and RULER~\citep{hsieh2024ruler}. We used the same advanced inference sparse methods, which also compute full attention scores using MHA and then perform sparsification based on that. The results are shown in Table~\ref{table:downstream_finetune}. DMA achieved the best average score, leading in GSM8K, MBPP, and LongBench, while remaining highly competitive in MMLU and RULER. NSA ranked first in BBH, with DMA closely following, while the full-attention MHA performed best in RULER but lagged behind DMA in average score. These results indicate that DMA's content-aware sparse mask effectively transfers even under longer context fine-tuning.

\paragraph{Extrapolated Content Retrieval.}
We further conducted an apples-to-apples comparison between MHA, NSA, and DMA using the needle-in-a-haystack task~\citep {gkamradt2023needleinahaystack} to evaluate the models' ability to retrieve information accurately from long texts. In this synthetic retrieval task, a random and information-rich sentence is inserted into a lengthy document, and the model needs to retrieve the needle from the haystack to answer the question. As shown in Figure~\ref{fig:needle_in_a_haystack}, as the context length increases, the advantage of DMA over NSA and MHA gradually expands. Notably, when the context length exceeds the pre-training sequence length, all three models exhibit a performance decline; however, the decrease in DMA's performance is significantly smaller than that of NSA and MHA, demonstrating stronger extrapolation capabilities and more effective retrieval of information in unseen length ranges. We speculate that trainable sparse attention inherently possesses stronger sequence length extrapolation. This experimental result has dual significance: on one hand, it validates DMA's intrinsic advantages in handling ultra-long documents, especially in practical application scenarios that require precise localization and extraction of key information; on the other hand, it reveals the structural advantages of DMA's content-aware dynamic mask mechanism in maintaining long-distance dependency modeling capabilities, even when the sequence length exceeds the pre-training range, thus maintaining relatively stable performance. This extrapolation capability is of great value for practical applications that require processing long documents.

Our comprehensive experimental results demonstrate the exceptional performance of Dynamic Mask Attention across various tasks and model scales. In scaling perplexity experiments, DMA consistently outperformed other attention variants across different parameter scales from 80M to 1.7B; in the multi-query associative recall task, DMA exhibited superior information retrieval capabilities and efficiency; in kernel implementations, DMA showed extremely high speedup ratios in various long-sequence application scenarios; in downstream benchmark evaluations, DMA models outperformed the original MHA and its various sparse variants on most tasks; in the needle-in-a-haystack task, DMA demonstrated significantly stronger length extrapolation capabilities. These results collectively validate the effectiveness of DMA as a sparse attention solution that simultaneously enhances computational efficiency and model performance.

\begin{figure}[!t]
    \centering
    \includegraphics[width=0.95\textwidth]{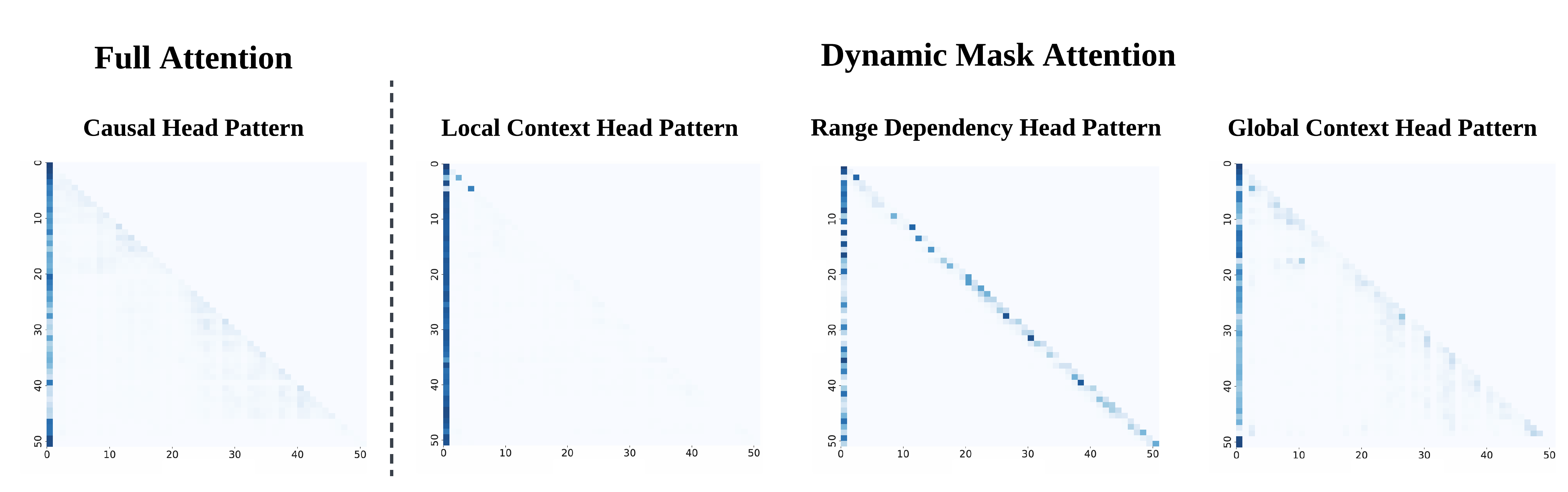} 
    \caption{
    \textbf{Weights Heatmaps of DMA}.
    The heatmaps show the attention weights of each head in the Dynamic Mask Attention mechanism, indicating which tokens each head focuses on. Full heatmaps can be found in Appendix~\ref{sec:appendix:attention_heatmaps}.
    }
    \label{fig:hatmaps}
\end{figure}
\vspace{-1.0em}
\begin{figure}[!t]
  \centering
  \includegraphics[width=0.95\linewidth]{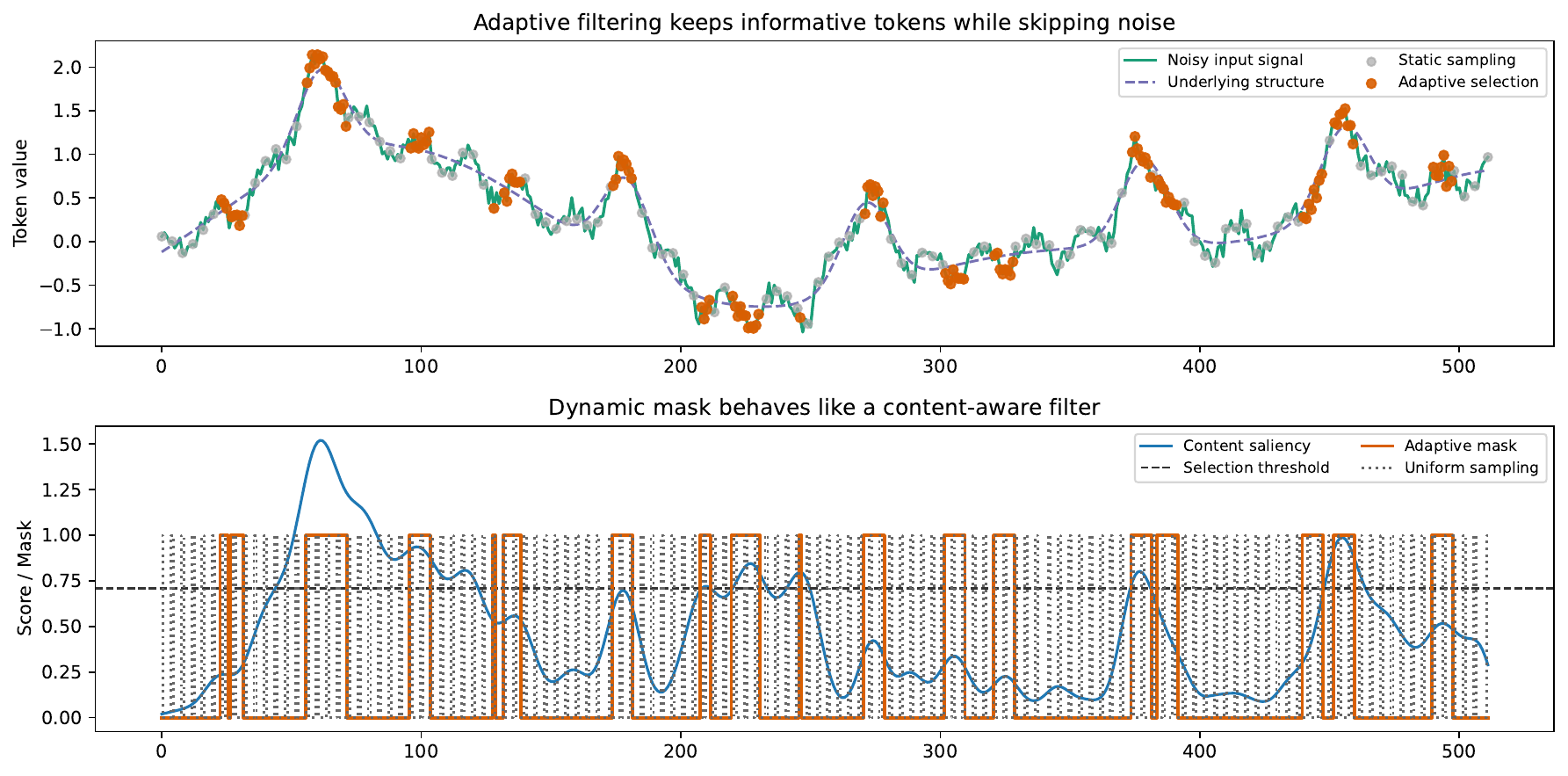}
  \caption{
  \textbf{Adaptive Filtering of DMA}.
  \textbf{Top}: the noisy token signal retains its underlying low-frequency structure while the adaptive mask focuses on informative tokens compared with uniform sampling.
  \textbf{Bottom}: the learned content-aware scores and the resulting mask illustrate how DMA allocates compute to relevant regions while skipping redundant context.
  }
  \label{fig:adaptive_filtering}
\end{figure}

\section{Analysis}
\label{sec:analysis}

In this section, we analyze Dynamic Mask Attention, highlighting its distinct advantages in handling long-range dependencies and providing dynamic context awareness.

\subsection{Head Specialization}
\label{subsec:head_specialization}

As shown in Figure~\ref{fig:hatmaps}~\ref{fig:full_heatmaps}, our analysis of the attention patterns learned by the model reveals how DMA creates content-aware sparse structures that adapt to different contextual needs. Unlike the uniform patterns of traditional attention mechanisms, each DMA attention head develops a unique sparse pattern: some heads focus on the most recent tokens to capture local context, while others attend to specific distant positions for long-range dependencies, and additional heads maintain broader context awareness for global understanding. This diversity allows the model to capture various types of dependencies simultaneously while maintaining computational efficiency, making efficient use of each subspace.

\textbf{Local Context Heads.} For example, Head 0, Head 1, Head 8, Head 10, and Head 11 tend to focus on the most recent tokens, forming a local band attention pattern. These heads are primarily responsible for capturing syntactic structures, phrase-level semantics, and local dependencies, which are particularly important for tasks requiring precise local context handling.

\textbf{Range Dependency Heads.} For example, Head 2, Head 3, Head 4, Head 5, and Head 14 demonstrate the ability to attend to specific distant tokens. These heads are specialized in capturing long-range semantic associations, such as resolving coreference issues or tracking complex storylines in lengthy documents. They can skip over large amounts of intermediate information and directly connect distant but semantically related parts, which is crucial for deep reasoning and contextual understanding.

\textbf{Global Context Heads.} For example, Head 6, Head 7, Head 9, Head 12, Head 13, and Head 15 exhibit a sparser but broader attention distribution, sampling key information from the entire sequence to form an overall perception of the global context. These heads function similarly to summarizers, responsible for integrating information from different parts to create a coherent global representation. This capability is crucial for tasks that require a comprehensive understanding of the entire input to make accurate predictions.

\textbf{Dynamic Adaptability.} The most significant advantage of DMA lies in its dynamism. These attention patterns are not static; they are dynamically generated based on the input content. This means the model can adjust its attention strategy in real-time, activating the most appropriate combination of heads when processing different tasks or text types. For example, when processing code, it might rely more on long-range dependency heads to track variable definitions and usages, whereas in a conversation, it might focus more on local context heads to understand the current exchange. This content-aware adaptability is the core advantage of DMA over static sparse attention methods.

This naturally occurring specialization is a direct result of the content-aware mask mechanism, enabling the model to effectively handle various types of dependencies while maintaining computational efficiency, achieving effective integration of multi-scale information. This hierarchical integration mechanism can effectively handle multi-level semantic structures in complex texts. It is worth noting that head specialization may also occur in traditional MHA, but the specialization patterns in DMA are more pronounced and functionally clearer, which may be a key reason for its superior performance across various tasks.

\subsection{Adaptive Filtering}

From the perspective of modern signal processing, Trainable Dynamic Mask Sparse Attention essentially performs dynamic downsampling of the input sequence through learnable adaptive filters, i.e., masks, retaining only key information components. This allows for efficient extraction of low-frequency dependencies in long-distance signals, such as text, while suppressing noise redundancy. The core logic is to treat long texts as noisy low-frequency signals, where the mask acts as an adaptive filter, and the dynamically selected retained tokens are equivalent to intelligent downsampling based on signal relevance.

\textbf{Learnable Content-aware Filters.} Unlike traditional sparse methods that rely on fixed patterns, DMA's mask is dynamically generated based on the input content, making it a content-aware adaptive filter. This filter learns to identify and amplify key components in the signal, i.e., important tokens, while attenuating or completely filtering out noise, i.e., irrelevant tokens. This mechanism ensures that computational resources are precisely allocated to the most critical information for the current task, effectively avoiding information loss due to excessively long contexts in needle-in-a-haystack problems.

\textbf{Multi-scale Signal Decomposition.} The different attention heads in DMA learn various sparse patterns, which can be viewed as a set of parallel adaptive filters, each responsible for capturing different scales or types of signal features. This multi-scale decomposition allows the model to build a comprehensive and hierarchical understanding of the input signal with extremely high efficiency.

Recasting sparse attention as an adaptive filtering problem, DMA offers a new perspective for understanding and optimizing long text processing. It achieves intelligent filtering of information through content awareness and multi-scale decomposition, ensuring that the model learns the optimal sparse strategies.

\section{Discussion}
\label{sec:discussion}

In this section, we discuss the core deficiencies of existing sparse attention methods, analyze how Dynamic Mask Attention addresses these issues, and explore its limitations and future development directions.

\subsection{Limitations of Existing Approaches}
\label{subsec:existing_limitations}

Existing sparse attention methods exhibit three critical deficiencies that limit their practical effectiveness:

\textbf{Post-hoc Sparsification Degradation.}
The performance degradation caused by post-hoc sparsification stems from the fundamental mismatch between existing methods and the optimization trajectory of pretrained models. As demonstrated by Chen et al.~\citep{chen2024magicpig}, retaining only the top 20\% of attention weights covers only 70\% of the total attention scores. This forced sparsification strategy compels models to deviate from the optimal parameter configurations learned on large-scale corpora. More critically, this approach causes irreversible damage to key structural components in pretrained models, such as retrieval heads and copy heads, as these specialized attention heads are misidentified as "unimportant" and pruned during inference. This structural destruction directly leads to significant performance degradation in tasks that require precise information retrieval and copying.

\textbf{Training-Inference Efficiency Gap.}
Most sparse attention methods optimize only for inference, neglecting training-phase computational demands. This creates bottlenecks across LLM development: pretraining on long documents, long-context fine-tuning, and reinforcement learning. Without effective training-time sparsity support, these crucial phases remain constrained by $O(n^2)$ computational complexity, limiting development of more capable long-context models.

\textbf{Non-differentiable Components and Inefficient Backpropagation.}
Non-differentiable components and inefficient backpropagation problems reveal the technical shortcomings of existing methods in terms of trainability. The discrete operations in methods like ClusterKV~\citep{liu2024clusterkv} and MagicPIG~\citep{chen2024magicpig} introduce discontinuities in computational graphs, which block gradient flow and hinder the learning of optimal sparse patterns. Even trainable methods like HashAttention~\citep{desai2024hashattention} suffer from memory access inefficiencies due to token-granular selection, which is incompatible with the contiguous memory access and block-wise computation requirements of efficient attention techniques, such as FlashAttention. Consequently, these implementations are forced to revert to naive implementations with low hardware utilization, significantly degrading training efficiency.

\subsection{How Dynamic Mask Attention Addresses Core Issues}
\label{subsec:dma_solutions}

Dynamic Mask Attention systematically addresses the aforementioned fundamental issues through three core innovations, achieving unified, efficient, and sparse computation for both training and inference phases.

\textbf{Native Trainable Sparsity.}
Native trainable sparsity is DMA's key innovation for addressing post-hoc sparsification issues. Unlike traditional methods, DMA embeds sparsity into the model architecture from the ground up, ensuring that sparse attention patterns are fully aligned with the model's optimization trajectory. Specifically, DMA retains complete, uncompressed KV caches $k = \operatorname{concat}([k_{1}, \ldots, k_{t}])$ and $v = \operatorname{concat}([v_{1}, \ldots, v_{t}])$, ensuring the original fidelity of historical information and precise recall capabilities, avoiding information bottlenecks that may arise from fixed-state compression in State Space Models. This comprehensive information retention mechanism enables DMA to precisely access any token in the historical context at any moment, without losing critical information due to lossy compression methods like Mamba. More importantly, DMA's sparsification occurs during the attention weight computation phase, rather than in post-training processing, ensuring that models do not deviate from pre-trained parameter configurations during sparsification, thereby protecting key structural components, such as retrieval heads and copy heads, from damage.

\textbf{Unified Training-Inference Architecture.}
The unified training-inference architecture eliminates the fundamental gap in training-inference efficiency that exists in existing methods. DMA's dynamic weight computation $\delta = \exp(\tau(v \Delta) \times A)$ uses identical sparsification strategies during both training and inference phases. This consistency ensures that models can learn optimal sparse patterns during training and seamlessly apply these patterns during inference. This unified architecture particularly benefits three critical stages of modern LLM development: the pretraining stage can efficiently process long document sequences; the long-context fine-tuning stage can adapt to specific task requirements; the reinforcement learning stage can effectively update attention weights through policy gradients. DMA reduces computational complexity from $O(n^2)$ to $O(n \cdot w)$, enabling the training of larger-scale long-context models.

\textbf{Fully Differentiable Design.}
The fully differentiable design ensures that DMA maintains gradient flow continuity throughout the entire computation process. The computation of dynamic mask weights $\delta$ is based entirely on differentiable operations, including linear transformations of value representations, non-negative activation functions $\tau(\cdot)$, and exponential functions, thereby avoiding gradient interruptions caused by discrete operations such as k-means clustering and SimHash. Although the mask generation process involves a top-k operation, it is not the core learning objective of DMA but merely a tool for sparse selection; thus, we only use this discrete operation in the forward pass. Moreover, the attention weight computation part is designed such that the gradients for masked positions should naturally be zero, so skipping computation and setting gradients to zero is the correct behavior. This design enables the model to learn optimal attention patterns that are sparse in an end-to-end manner, dynamically adjusting which historical positions are most critical for current reasoning, thereby achieving truly content-aware, selective computation. Additionally, each head in a multi-head attention mechanism can independently generate different sparse patterns, thereby the representational capabilities of the multi-head architecture by focusing on different information segments in distinct subspaces.

\subsection{Limitations and Future Works}
\label{subsec:limitations_future}

Despite Dynamic Mask Attention's significant progress in addressing the core issues of existing methods, several limitations remain that warrant further exploration and improvement in future work.

\textbf{Adaptive Window Size Selection.}
Adaptive window size selection is the primary challenge facing DMA. While the current fixed window size design provides predictable computational complexity, it may not optimally adapt to the dynamic demands of different tasks and contexts. For instance, code generation tasks may require larger windows to capture long-range structural dependencies, while simple question-answering tasks may only need smaller windows. Future research directions include developing adaptive window size selection mechanisms based on task complexity, sequence length, and content features, potentially through reinforcement learning or meta-learning approaches to dynamically optimize window parameters. Alternatively, designing hierarchical multi-scale attention structures can be considered to capture dependencies across different ranges simultaneously.

\textbf{Position Encoding Enhancement.}
Our needle-in-a-haystack experiments revealed an intriguing phenomenon: trainable sparse attention mechanisms, such as DMA, exhibit stronger length extrapolation capabilities compared to dense attention when context lengths exceed the pretraining bounds. This finding suggests that the fundamental bottleneck for extrapolation may lie in the position encoding method rather than the attention mechanism itself. Current RoPE-based position encodings struggle with out-of-distribution sequence lengths, but DMA's dynamic sampling architecture offers a potential alternative pathway for encoding positional information. Specifically, the zero-order hold sampling values that are added as attention biases can be explored to explicitly incorporate positional information into these sampling values, potentially replacing or complementing RoPE to create a more extrapolation-friendly encoding scheme.
Such an approach might leverage the inherent advantages of sparse attention's selective computation to create position representations that scale more naturally to unseen lengths. This direction could help address one of the most persistent challenges in long-context modeling: maintaining consistent positional understanding across arbitrary sequence lengths without requiring length-specific fine-tuning.

\textbf{Multi-Modal Extension.}
Multi-modal extension represents an essential direction for DMA development. The current DMA design is primarily optimized for text sequences; however, modern AI systems increasingly require processing mixed inputs of text, images, audio, and video. Attention sparsity in multi-modal scenarios exhibits more complex patterns: interactions between different modalities may require different attention distributions, temporally aligned multi-modal information may need synchronized attention mechanisms, and modality-specific long-range dependencies may require specialized sparse patterns. Future research can explore modality-aware dynamic mask generation, coordination mechanisms for cross-modal attention weights, and specialized sparse pattern designs for different modal characteristics.

\textbf{Integration with Modern Frameworks.}
Seamless integration of DMA into mainstream deep learning frameworks, such as PyTorch and the Hugging Face Transformers library, is crucial for its widespread adoption and practical impact. This requires developing a user-friendly and highly optimized implementation that can be easily incorporated into existing model architectures and training pipelines. A key aspect of this integration is the development of efficient, low-level kernels, potentially using Triton or CUDA, to ensure that the performance benefits of DMA are fully realized on modern hardware. Providing a plug-and-play module compatible with the Hugging Face Transformers library would significantly lower the barrier for researchers and practitioners to apply DMA to their models and tasks, thereby fostering further innovation and comparative studies in the field of sparse attention.

\section{Conclusion}
\label{sec:conclusion}

In this paper, we introduced Dynamic Mask Attention, a novel trainable sparse attention mechanism that effectively addresses the key challenges in long-context modeling for large language models. By integrating content-aware dynamic sparse masks with position-aware sparse attention weight computations, Dynamic Mask Attention successfully balances computational efficiency while preserving the ability to retrieve information from long contexts precisely.

Our approach makes several key contributions to the field of efficient attention mechanisms. First, Dynamic Mask Attention achieves computational efficiency comparable to sliding window attention while maintaining the information retrieval capabilities of full attention by retaining a complete, uncompressed key-value cache. Second, by dynamically generating attention masks from value representations, our method enables models to learn which tokens are relevant to the current reasoning process, effectively leveraging both content-aware and position-aware sparsity patterns inherent in language modeling tasks. Third, our specialized hardware-optimized kernel for Dynamic Mask Attention efficiently handles sparse mask regions, translating theoretical computational gains into practical speed improvements.

The comprehensive experimental evaluation demonstrates that Dynamic Mask Attention consistently outperforms existing attention mechanisms across various scales and tasks. In scaling law studies, Dynamic Mask Attention exhibited superior perplexity compared to other attention variants. On challenging tasks like multi-query associative recall, Dynamic Mask Attention demonstrated both effectiveness in information retrieval and computational efficiency. Most significantly, our 1.7B parameter model with Dynamic Mask Attention outperformed the vanilla attention counterpart on standard benchmarks and showed remarkably stronger extrapolation capabilities on the needle-in-a-haystack task when context lengths exceeded the pre-training sequence length.

Dynamic Mask Attention represents a significant step forward in developing efficient and effective attention mechanisms for long-context modeling. By maintaining the full expressive power of attention while reducing computational complexity, our approach enables the development of more capable language models that can effectively process lengthy documents, complex reasoning chains, and rich contextual information. This capability is particularly valuable for applications requiring deep reasoning, code generation, and multi-turn autonomous agents.

Future work could explore adaptive window size selection based on content complexity, create more extrapolation-friendly positional encoding schemes, extend it to multimodal contexts, and develop further theoretical analyses of its properties. We believe that Dynamic Mask Attention provides a promising direction for future research in efficient transformer architectures and will facilitate the development of more powerful and computationally efficient language models.

\subsubsection*{Acknowledgments}
We would like to express our gratitude to the OpenSeek project team at Beijing Academy of Artificial Intelligence for their support in developing the hardware kernels. We would also like to thank Professor Albert Gu from Carnegie Mellon University for his valuable guidance and suggestions in connecting the concepts of state-space and self-attention. Additionally, we extend our thanks to all the friends in the community who provided feedback and suggestions, your support has been instrumental in the continuous improvement of this work.

\newpage

\printbibliography

\newpage

\appendix

\onecolumn

\section{Dynamic Mask Attention Implementation}
\label{sec:appendix:dma_implementation}

The following listing provides a sample implementation of the Dynamic Mask Attention algorithm in PyTorch, as described in Section~\ref{sec:methods}.

\lstdefinelanguage{PythonArXiv}[]{Python}{
  morekeywords={self,torch,nn,einsum,arange,where,topk,matmul,softmax},
  sensitive=true
}
\lstset{
  language=PythonArXiv,
  basicstyle=\ttfamily\small,
  keywordstyle=\bfseries, 
  commentstyle=\itshape,
  stringstyle=\ttfamily,
  showstringspaces=false,
  breaklines=true,
  columns=fullflexible,
  keepspaces=true,
  frame=single,
  framerule=0.3pt,
}

\begin{lstlisting}[caption={Dynamic Mask Attention implementation in PyTorch},label={lst:dynamic_mask_attn_code}]
def dynamic_mask_attention(h_t, position_embeddings, causal_m, past_key_value,
    W_Q, W_K, W_V, W_dt, A, W_O,
    num_heads, scaling, keep_window_size):  
    input_shape = h_t.shape[:-1]                            # [b, q_len]
    hidden_shape = (*input_shape, -1, h_t.shape[-1] // num_heads)
    # linear projections
    q_t = W_Q(h_t).view(hidden_shape).transpose(1, 2)       # [b, n_h, q_len, d_h]
    k_t = W_K(h_t).view(hidden_shape).transpose(1, 2)       # [b, n_h, q_len, d_h]
    v_t = W_V(h_t).view(hidden_shape).transpose(1, 2)       # [b, n_h, q_len, d_h]
    o_t = torch.zeros_like(q_t)                             # [b, n_h, q_len, d_h]
    # apply rotary position embeddings
    q_t, k_t = apply_rotary_pos_emb(q_t, k_t, *position_embeddings)
    # concatenate past key and value states
    k, v = past_key_value.update(k_t, v_t)                  # [b, n_h, k_len, d_h]
    # calculate dynamic mask
    dt = W_dt(v.transpose(1, 2).reshape(v.shape[0], v.shape[-2], -1))   # [b, k_len, n_h]
    dt = torch.exp(A * F.softplus(dt)).transpose(-1, -2)                # [b, n_h, k_len]
    m_t = dt[:, :, None, :].expand(-1, -1, h_t.shape[1], -1)            # [b, n_h, q_len, k_len]
    active_m = torch.zeros_like(m_t)
    m_t = m_t.masked_fill(causal_m != 0, -float('inf'))
    topk_indices = torch.topk(m_t, keep_window_size, dim=-1, sorted=False).indices
    active_m = active_m.scatter(-1, topk_indices, 1.0)
    m_t = m_t.masked_fill(active_m == 0.0, -float('inf'))
    # calculate sparse attention weight
    for b_idx in range(hidden_shape[0]):                                            # b
        for h_idx in range(num_heads):                                              # n_h
            for q_idx in range(hidden_shape[1]):                                    # q_len
                q_elem = q_t[b_idx, h_idx, q_idx, :]                                # [d_h]
                indices = topk_indices[b_idx, h_idx, q_idx]                         # [w]
                k_vecs = k[b_idx, h_idx, indices, :]                                # [w, d_h]
                v_vecs = v[b_idx, h_idx, indices, :]                                # [w, d_h]
                a_elem = torch.sum(q_elem.unsqueeze(0) * k_vecs, dim=-1)            # [w]
                a_elem = a_elem * scaling + m_t[b_idx, h_idx, q_idx, indices]
                a_elem = F.softmax(a_elem, dim=-1)
                o_elem = torch.sum(a_elem.unsqueeze(1) * v_vecs, dim=0)             # [d_h]
                o_t[b_idx, h_idx, q_idx, :] = o_elem
    o_t = o_t.transpose(1, 2).contiguous()                  # [b, q_len, n_h, d_h]
    h_t = W_O(o_t.view(*input_shape, -1))                   # [b, q_len, d_model]
    return h_t
\end{lstlisting}

The implementation demonstrates the core computational flow of the Dynamic Mask Attention mechanism. First, the query, key, and value matrices are computed through linear projections, followed by the application of rotary position embeddings. The core innovation of the algorithm is then reflected in the dynamic mask generation process: dynamic weights $\delta$ are calculated from the value vectors, and a sparse mask is generated using the topk operation, retaining only the most relevant $w$ key-value pairs. Finally, in the sparse attention computation phase, the algorithm computes attention weights only for the selected key-value pairs, significantly reducing computational complexity. In actual kernel implementations, it is possible to check if there are any active tokens in the MMA block; if not, the computation for that block can be skipped.

\newpage
\section{Experiment Setup}
\label{sec:appendix:experiment_setup}

To make the comparison between attention variants fair and reproducible, we standardize the model, data pipeline, optimization, and evaluation, only changing the attention module and its related hyperparameters. All experiments were conducted using the open-source PyTorch images~\citep{nv2022pytorch} and the Transformers framework~\citep{wolf-etal-2020-transformers}. We use SmolLMCorpus~\citep{benallal2024smollmcorpus} as training data. For evaluation frameworks, we utilized the LM evaluation harness~\citep{eval-harness} from EleutherAI for perplexity tasks, and the lighteval~\citep{Clémentine2023lighteval} from HuggingFace for downstream tasks. Table~\ref{tab:scaling_laws_configurations} lists the model sizes and key hyperparameters used at each scale. We keep the depth, width, and number of heads consistent across variants under the same parameter budget, only changing the specific parameters of the attention variants, with parameter symbols consistent with those in the original papers of the attention variants.

We summarize the meaning of the columns in Table~\ref{tab:scaling_laws_configurations} and clarify which hyperparameters are used by each attention variant.

\begin{itemize}
    \item \textbf{Params}: total number of model parameters.
    \item \textbf{Steps}: training steps.
    \item \textbf{Batch}: tokens per step.
    \item \textbf{LR}: peak learning rate.
    \item $n_{layers}$: number of Transformer layers.
    \item $d_{model}$: model hidden size.
    \item $n_h$: number of attention heads.
    \item $n_{h_{kv}}$: number of KV heads. In all our configurations we set $n_{h_{kv}} = n_h/2$.
\end{itemize}

\begin{itemize}
    \item \textbf{MHA} (full attention): standard scaled dot-product attention. Sparse-specific columns ($w$, $d_c$, $B$, $B'$, $k$) are not used.
    \item \textbf{SWA} (sliding-window attention): $w$ is the sliding window size; other sparse-specific columns are not used.
    \item \textbf{MLA} (multi-head latent attention): $d_c$ is the latent/compression dimension; other columns are not used.
    \item \textbf{NSA} (native sparse attention): $d_c$ is the compression dimension, $w$ is the sliding window size, $B$ is the compressing block size, $B'$ is the selection block size, and $k$ is the num selected blocks. All settings following the original paper.
    \item \textbf{DMA} (dynamic mask attention): $w$ is the per-head top-$w$ kept keys for dynamic masks; other sparse-specific columns are not used.
\end{itemize}

\begin{table}[!t]
    \centering
    \small
    \caption{
    \textbf{Self-Attention Variants Scaling Laws Configurations}.
    The model and hyperparameter configurations used in our self-attention variants scaling laws experiments.
    }
    \label{tab:scaling_laws_configurations}
    \resizebox{\linewidth}{!}
    {
    \begin{tabular}{@{}lccccccccccccc@{}}
    \toprule
    \sc{Algos} & \sc{Params} & \sc{Steps} & \sc{Batch} & \sc{LR} & $n_{layers}$ & $d_{model}$ & $n_{h}$ & $n_{h_{kv}}$ & \sc{$w$} & \sc{$d_c$} & \sc{$B$} & \sc{$B'$} & \sc{$k$} \\
    \midrule
    MHA & $\approx$ 80M & 13,500 & 0.128M tokens & 3e-3 & 12 & 768 & 6 & 3 & - & - & - & - & - \\
    SWA & $\approx$ 80M & 13,500 & 0.128M tokens & 3e-3 & 12 & 768 & 6 & 3 & 1024 & - & - & - & - \\
    MLA & $\approx$ 80M & 13,500 & 0.128M tokens & 3e-3 & 12 & 768 & 6 & 3 & - & 192 & - & - & - \\
    NSA & $\approx$ 80M & 13,500 & 0.128M tokens & 3e-3 & 12 & 768 & 6 & 3 & 512 & 192 & 32 & 64 & 16 \\
    DMA & $\approx$ 80M & 13,500 & 0.128M tokens & 3e-3 & 12 & 768 & 6 & 3 & 1024 & - & - & - & - \\
    \midrule
    MHA & $\approx$ 200M & 20,800 & 0.192M tokens & 2e-3 & 16 & 1024 & 8 & 4 & - & - & - & - & - \\
    SWA & $\approx$ 200M & 20,800 & 0.192M tokens & 2e-3 & 16 & 1024 & 8 & 4 & 1024 & - & - & - & - \\
    MLA & $\approx$ 200M & 20,800 & 0.192M tokens & 2e-3 & 16 & 1024 & 8 & 4 & - & 256 & - & - & - \\
    NSA & $\approx$ 200M & 20,800 & 0.192M tokens & 2e-3 & 16 & 1024 & 8 & 4 & 512 & 192 & 32 & 64 & 16 \\
    DMA & $\approx$ 200M & 20,800 & 0.192M tokens & 2e-3 & 16 & 1024 & 8 & 4 & 1024 & - & - & - & - \\
    \midrule
    MHA & $\approx$ 680M & 35,000 & 0.392M tokens & 1e-3 & 24 & 1536 & 12 & 6 & - & - & - & - & - \\
    NSA & $\approx$ 680M & 35,000 & 0.392M tokens & 1e-3 & 24 & 1536 & 12 & 6 & 512 & 192 & 32 & 64 & 16 \\
    DMA & $\approx$ 680M & 35,000 & 0.392M tokens & 1e-3 & 24 & 1536 & 12 & 6 & 1024 & - & - & - & - \\
    \midrule
    MHA & $\approx$ 1.7B & 40,000 & 1M tokens & 1e-3 & 28 & 2048 & 16 & 8 & - & -  & - & - & - \\
    NSA & $\approx$ 1.7B & 40,000 & 1M tokens & 1e-3 & 28 & 2048 & 16 & 8 & 512 & 256 & 32 & 64 & 16 \\
    DMA & $\approx$ 1.7B & 40,000 & 1M tokens & 1e-3 & 28 & 2048 & 16 & 8 & 2048 & - & - & - & - \\
    \bottomrule
    \end{tabular}
    }
\end{table}

\footnotetext[1]{The implementation code for MHA is available at \url{https://github.com/Dao-AILab/flash-attention}.}
\footnotetext[2]{The implementation code for SWA is available at \url{https://github.com/Dao-AILab/flash-attention}.}
\footnotetext[3]{The implementation code for MLA is available at \url{https://github.com/deepseek-ai/FlashMLA}.}
\footnotetext[4]{The implementation code for NSA is available at \url{https://github.com/lucidrains/native-sparse-attention-pytorch}.}

\begin{table}[!h]
    \centering
    \small
    \caption{
    \textbf{Speed Benchmark Configurations}.
    The common settings for running time curves and the sparse hyperparameters for each method.
    }
    \label{tab:speed_benchmark_configs}
    \begin{tabular}{@{}lcccccccccccccc@{}}
    \toprule
    \sc{Algo} & warmups & runs & $n_h$ & $n_{h_{kv}}$ & $d_h$ & $w$ & $d_c$ & $B$ & $B'$ & $k$ & precision \\
    \midrule
    MHA\footnotemark[1] & 3 & 1,000 & 32 & 8 & 128 & -    & -   & -  & -  & - & bf16 \\
    SWA\footnotemark[2] & 3 & 1,000 & 32 & 8 & 128 & 1024 & -   & -  & -  & - & bf16 \\
    MLA\footnotemark[3] & 3 & 1,000 & 32 & 8 & 128 & -    & 192 & -  & -  & - & bf16 \\
    NSA\footnotemark[4] & 3 & 1,000 & 32 & 8 & 128 & 512  & 256 & 32 & 64 & 16 & bf16\\
    DMA & 3 & 1,000 & 32 & 8 & 128 & 1024 & -   & -  & -  & - & bf16 \\
    \bottomrule
    \end{tabular}
\end{table}

\clearpage
\newpage
\section{Attention Heatmaps}
\label{sec:appendix:attention_heatmaps}

\begin{figure}[!h]
    \centering
    \includegraphics[width=0.95\textwidth]{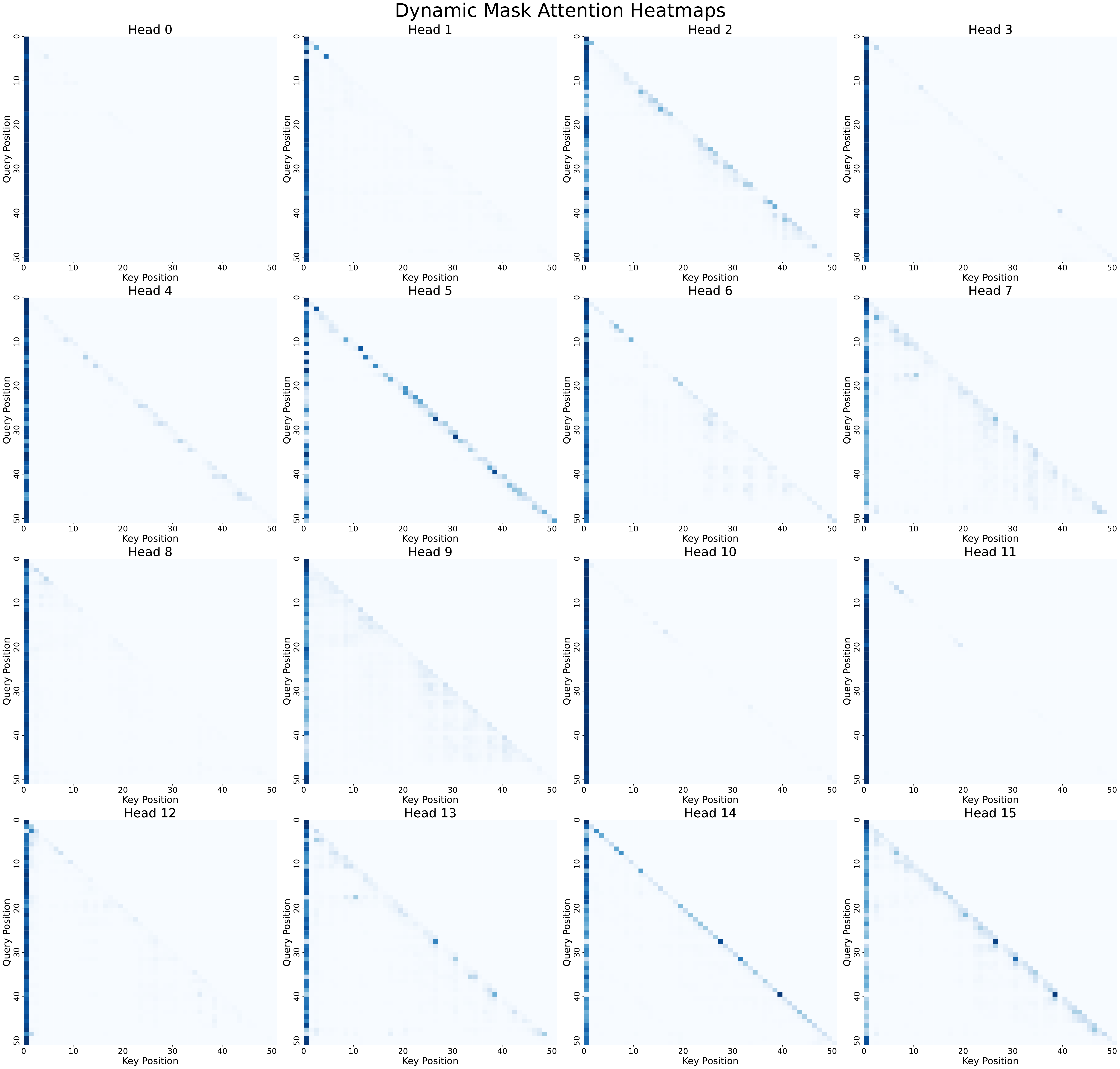} 
    \caption{
    \textbf{Full Heatmaps of Dynamic Mask Attention}.
    The heatmaps show the attention weights of each head in the Dynamic Mask Attention mechanism, indicating which tokens each head focuses on.
    }
    \label{fig:full_heatmaps}
\end{figure}

\end{document}